%% file: doc.tex
\documentclass{article}  

\usepackage{booktabs}
\usepackage{hyperref}

\usepackage[accepted]{mlsys2020}

\newcommand{\npapers}{\text{81 }}

\input{setup.tex}

\mlsystitlerunning{What is the State of Neural Network Pruning?}

\begin{document}

\twocolumn[
\mlsystitle{What is the State of Neural Network Pruning?}

\mlsyssetsymbol{equal}{*}
\begin{mlsysauthorlist}
\mlsysauthor{Davis Blalock}{equal,csail}
\mlsysauthor{Jose Javier Gonzalez Ortiz}{equal,csail}
\mlsysauthor{Jonathan Frankle}{csail}
\mlsysauthor{John Guttag}{csail}
\end{mlsysauthorlist}

\mlsysaffiliation{csail}{MIT CSAIL, Cambridge, MA, USA}
\mlsyscorrespondingauthor{Davis Blalock}{dblalock@mit.edu}

\mlsyskeywords{Deep Learning, Pruning}

\vskip 0.2in

\begin{abstract}

Neural network pruning---the task of reducing the size of a network by removing parameters---has been the subject of a great deal of work in recent years. We provide a meta-analysis of the literature, including an overview of approaches to pruning and consistent findings in the literature. After aggregating results across \npapers papers and pruning hundreds of models in controlled conditions, our clearest finding is that the community suffers from a lack of standardized benchmarks and metrics.
This deficiency is substantial enough that it is hard to compare pruning techniques to one another or determine how much progress the field has made over the past three decades.
To address this situation, we identify issues with current practices, suggest concrete remedies, and introduce ShrinkBench, an open-source framework to facilitate standardized evaluations of pruning methods. We use ShrinkBench to compare various pruning techniques and show that its comprehensive evaluation can prevent common pitfalls when comparing pruning methods.

\end{abstract}
]  


\printAffiliationsAndNotice{\mlsysEqualContribution} 

\section{Introduction} \label{sec:intro}
\vspace{-.75mm}

\input{intro.tex}
\section{Overview of Pruning}

\input{overview_v2.tex}

\vspace{-2mm}
\section{Lessons from the Literature} \label{sec:lessons}
\vspace{-.5mm}

\input{lessons2.tex}
\section{Missing Controlled Comparisons}

\input{is_problem.tex}

\section{Further Barriers to Comparison}

\input{why_problem.tex}

\section{Summary and Recommendations}

\input{recommendations.tex}

\section{ShrinkBench} \label{sec:bench}

\input{shrinkbench_v2.tex}

\vspace{-1mm}
\section{Conclusion}

Considering the enormous interest in neural network pruning over the past decade, it seems natural to ask simple questions about the relative efficacy of different pruning techniques.
Although a few basic findings are shared across the literature, missing baselines and inconsistent experimental settings make it impossible to assess the state of the art or confidently compare the dozens of techniques proposed in recent years.
After carefully studying the literature and enumerating numerous areas of incomparability and confusion, we suggest concrete remedies in the form of a list of best practices and an open-source library---ShrinkBench---to help future research endeavors to produce the kinds of results that will harmonize the literature and make our motivating questions easier to answer. Furthermore, ShrinkBench results on various pruning techniques evidence the need for standardized experiments when evaluating neural network pruning methods.

\section*{Acknowledgements}

We thank Luigi Celona for providing the data used in \cite{luigi} and Vivienne Sze for helpful discussion. This research was supported by the Qualcomm Innovation Fellowship, the ``la Caixa'' Foundation Fellowship, Quanta Computer, and Wistron Corporation.



\bibliographystyle{mlsys2020}
\bibliography{combined}

\clearpage
\newpage  
\appendix


\appendix

\input{appendix_corpus}
\input{appendix_checklist}
\input{appendix_framework}
\input{appendix_results}

\end{document}

%% file: setup.tex
\usepackage{bbm}  

\usepackage{amsmath}          
\usepackage{amssymb} 			    
\usepackage{textcomp}

\usepackage{url}              

\usepackage{array}            
\usepackage{tabularx}         
\usepackage{colortbl}
\newcolumntype{Y}{>{\centering\arraybackslash}X}	

\usepackage{mathtools}          

\usepackage{pbox}   

\usepackage{setspace}

\usepackage{graphicx}
\usepackage[space]{grffile}   
\graphicspath{{./figs/}}
\DeclareGraphicsExtensions{.pdf,.jpeg,.jpg,.png}

\usepackage[font={bf}]{caption}
\usepackage{outlines}
\usepackage{enumitem}
\newcommand{\ItemSpacing}{0mm}
\newcommand{\ParSpacing}{0mm}
\setenumerate[1]{itemsep={\ItemSpacing},parsep={\ParSpacing},label=\arabic*.}
\setenumerate[2]{itemsep={\ItemSpacing},parsep={\ParSpacing}}

\newcommand\eps\varepsilon

\renewcommand\inf\infty



%% file: intro.tex

Much of the progress in machine learning in the past decade has been a result of deep neural networks. Many of these networks, particularly those that perform the best \cite{gpipe}, require enormous amounts of computation and memory. These requirements not only increase infrastructure costs, but also make deployment of networks to resource-constrained environments such as mobile phones or smart devices challenging \cite{learning-both, szeEfficient, sze-energy-aware}.

One popular approach for reducing these resource requirements at test time is neural network \textit{pruning}, which entails systematically removing parameters from an existing network. Typically, the initial network is large and accurate, and the goal is to produce a smaller network with similar accuracy. Pruning has been used since the late 1980s \cite{janowsky_pruning_1989, mozer_skeletonization:_1989, mozer_using_1989, karnin_simple_1990}, but has seen an explosion of interest in the past decade thanks to the rise of deep neural networks.

For this study, we surveyed 81 recent papers on pruning in the hopes of extracting practical lessons for the broader community.
For example: which technique achieves the best accuracy/efficiency tradeoff?
Are there strategies that work best on specific architectures or datasets?
Which high-level design choices are most effective?

There are indeed several consistent results: pruning parameters based on their magnitudes substantially compresses networks without reducing accuracy, and many pruning methods outperform random pruning.
However, our central finding is that the state of the literature is such that our motivating questions are impossible to answer.
Few papers compare to one another, and methodologies are so inconsistent between papers that we could not make these comparisons ourselves.
For example, a quarter of papers compare to no other pruning method, half of papers compare to at most one other method, and
dozens of methods have never been compared to by any subsequent work.
In addition, no dataset/network pair appears in even a third of papers, evaluation metrics differ widely, and hyperparameters and other counfounders vary or are left unspecified.
Most of these issues stem from the absence of standard datasets, networks, metrics, and experimental practices. 
To help enable more comparable pruning research,
we identify specific impediments and pitfalls, recommend best practices, and introduce ShrinkBench, a library for standardized evaluation of pruning. ShrinkBench makes it easy to adhere to the best practices we identify, largely by providing a standardized collection of pruning primitives, models, datasets, and training routines.

Our contributions are as follows:%
\begin{enumerate}[leftmargin=5mm]
    \itemsep1pt
    \vspace{-2mm}
    \item A meta-analysis of the neural network pruning literature based on comprehensively aggregating reported results from \npapers papers.
    \item A catalog of problems in the literature and best practices for avoiding them. These insights derive from analyzing existing work and pruning hundreds of models.
    \item ShrinkBench, an open-source library for evaluating neural network pruning methods available at \\{ \url{https://github.com/jjgo/shrinkbench}}.
\end{enumerate}

%% file: overview_v2.tex
\renewcommand{\algorithmicrequire}{\textbf{Input:}}
\label{sec:overview}
Before proceeding, we first offer some background on neural network pruning and a high-level overview of how existing pruning methods typically work.

\subsection{Definitions}

We define a neural network \emph{architecture} as a function family $f(x; \cdot)$.
The architecture consists of the configuration of the network's parameters and the sets of operations it uses to produce outputs from inputs, including the arrangement of parameters into convolutions, activation functions, pooling, batch normalization, etc.
Example architectures include AlexNet and ResNet-56.
We define a neural network \emph{model} as a particular parameterization of an architecture, i.e., $f(x; W)$ for specific parameters $W$.
Neural network \emph{pruning} entails taking as input a model $f(x; W)$ and producing a new model $f(x; M \odot W')$. Here $W'$ is set of parameters that may be different from $W$, $M \in \{0, 1\}^{|W'|}$ is a binary mask that fixes certain parameters to $0$, and $\odot$ is the elementwise product operator.
In practice, rather than using an explicit mask, pruned parameters of $W$ are fixed to zero or removed entirely.

\subsection{High-Level Algorithm}

There are many methods of producing a pruned model $f(x; M \odot W')$ from an initially untrained model $f(x; W_0)$, where $W_0$ is sampled from an initialization distribution $\mathcal{D}$.
Nearly all neural network pruning strategies in our survey derive from Algorithm \ref{alg:prune-after-training} \cite{learning-both}.
In this algorithm, the network is first trained to convergence.
Afterwards, each parameter or structural element in the network is issued a score, and the network is pruned based on these scores.
Pruning reduces the accuracy of the network, so it is trained further (known as \emph{fine-tuning}) to recover.
The process of pruning and fine-tuning is often iterated several times, gradually reducing the network's size.

Many papers propose slight variations of this algorithm.
For example, some papers prune periodically during training \cite{google-state-of-sparsity} or even at initialization \cite{snip}.
Others modify the network to explicitly include additional parameters that encourage sparsity and serve as a basis for scoring the network after training \cite{sparse-variational-dropout}.

\begin{algorithm}[h]
\caption{Pruning and Fine-Tuning}
\label{alg:prune-after-training}
\begin{algorithmic}[1]
\REQUIRE $N$, the number of iterations of pruning, and \\ \hspace{1.5em}$X$, the dataset on which to train and fine-tune
    \STATE $W \gets initialize()$
    \STATE $W \gets trainToConvergence(f(X; W))$

    \STATE $M \gets 1^{|W|}$
    \FOR{$i$ \text{ }in $1$ to $N$}%
        \STATE $M \gets prune(M, score(W))$%
        \STATE $W \gets fineTune(f(X; M \odot W))$%
    \ENDFOR
    \STATE \textbf{return} $M, W$
\end{algorithmic}
\end{algorithm}

\subsection{Differences Betweeen Pruning Methods}

Within the framework of Algorithm \ref{alg:prune-after-training}, pruning methods vary primarily in their choices regarding sparsity structure, scoring, scheduling, and fine-tuning.

\textbf{Structure.} Some methods prune individual parameters (\emph{unstructured pruning}). Doing so produces a sparse neural network, which---although smaller in terms of parameter-count---may not be arranged in a fashion conducive to speedups using modern libraries and hardware.
Other methods consider parameters in groups (\emph{structured pruning}), removing entire neurons, filters, or channels to exploit hardware and software optimized for dense computation \cite{pruning-filters, channel-lasso-lstsq}.

\textbf{Scoring.}
It is common to score parameters based on their absolute values, trained importance coefficients, or contributions to network activations or gradients. 
Some pruning methods compare scores locally, pruning a fraction of the parameters with the lowest scores within each structural subcomponent of the network (e.g., layers) \cite{learning-both}.
Others consider scores globally, comparing scores to one another irrespective of the part of the network in which the parameter resides \cite{snip, lottery-ticket}.


\textbf{Scheduling.}
Pruning methods differ in the amount of the network to prune at each step.
Some methods prune all desired weights at once in a single step \cite{rethinking-net-pruning}.
Others prune a fixed fraction of the network iteratively over several steps \cite{learning-both} or vary the rate of pruning according to a more complex function \cite{google-state-of-sparsity}.

\textbf{Fine-tuning.}
For methods that involve fine-tuning, it is most common to continue to train the network using the trained weights from before pruning.
Alternative proposals include rewinding the network to  an earlier state \cite{lottery-ticket-followup} and reinitializing the network entirely \cite{rethinking-net-pruning}.

\subsection{Evaluating Pruning}

Pruning can accomplish many different goals, including reducing the storage footprint of the neural network, the computational cost of inference, the energy requirements of inference, etc.
Each of these goals favors different design choices and requires different evaluation metrics.
For example, when reducing the storage footprint of the network, all parameters can be treated equally, meaning one should evaluate the overall compression ratio achieved by pruning.
However, when reducing the computational cost of inference, different parameters may have different impacts.
For instance, in convolutional layers, filters applied to spatially larger inputs are associated with more computation than those applied to smaller inputs.

Regardless of the goal, pruning imposes a tradeoff between model efficiency and quality, with pruning increasing the former while (typically) decreasing the latter. This means that a pruning method is best characterized not by a single model it has pruned, but by a family of models corresponding to different points on the efficiency-quality curve.
To quantify efficiency, most papers report at least one of two metrics. The first is the number of multiply-adds (often referred to as FLOPs) required to perform inference with the pruned network. The second is the fraction of parameters pruned. To measure quality, nearly all papers report changes in Top-1 or Top-5 image classification accuracy.

As others have noted \cite{lempitsky-cp-decomp, perforated-cnns, bayesian-compression, sze-energy-aware, learning-both, samsung-vbmf-tucker, ssl, thinet-channel-norms, amc-automl-han}, these metrics are far from perfect. Parameter and FLOP counts are a loose proxy for real-world latency, throughout, memory usage, and power consumption. 
Similarly, image classification is only one of the countless tasks to which neural networks have been applied. However, because the overwhelming majority of papers in our corpus focus on these metrics, our meta-analysis necessarily does as well.




%% file: lessons2.tex

After aggregating results from a corpus of \npapers papers, we identified a number of consistent findings. In this section, we provide an overview of our corpus and then discuss these findings.

\vspace{-.75mm}
\subsection{Papers Used in Our Analysis}
\vspace{-.25mm}

Our corpus consists of 79 pruning papers published since 2010 and two classic papers \cite{optimal-brain-damage, optimal-brain-surgeon} that have been compared to by a number of recent methods. We selected these papers by identifying popular papers in the literature and what cites them, systematically searching through conference proceedings, and tracing the directed graph of comparisons between pruning papers. This last procedure results in the property that, barring oversights on our part, there is no pruning paper in our corpus that compares to any pruning paper outside of our corpus. Additional details about our corpus and its construction can be found in Appendix~\ref{sec:corpus}.


\vspace{-1mm}
\subsection{How Effective is Pruning?}
\vspace{-.25mm}

One of the clearest findings about pruning is that it works. More precisely, there are various methods that can significantly compress models with little or no loss of accuracy. In fact, for small amounts of compression, pruning can sometimes \textit{increase} accuracy \cite{learning-both, spectral-pruning}. This basic finding has been replicated in a large fraction of the papers in our corpus. 

Along the same lines, it has been repeatedly shown that, at least for large amounts of pruning, many pruning methods outperform random pruning \cite{nisp, google-state-of-sparsity, lottery-ticket-followup, divnet, apple-pfa, channel-lasso-lstsq}. Interestingly, this does not always hold for small amounts of pruning \cite{lottery-transfer}. Similarly, pruning all layers uniformly tends to perform worse than intelligently allocating parameters to different layers \cite{google-state-of-sparsity, learning-both, pruning-filters, nvidia-taylor-pruning, thinet-channel-norms} or pruning globally \cite{snip, lottery-ticket}. Lastly, when holding the number of fine-tuning iterations constant, many methods produce pruned models that outperform retraining from scratch with the same sparsity pattern \cite{zhang-accel-very-deep, nisp, bayesian-compression, channel-lasso-lstsq, thinet-channel-norms, lottery-ticket} (at least with a large enough amount of pruning \cite{apple-pfa}). Retraining from scratch in this context means training a fresh, randomly-initialized model with all weights clamped to zero throughout training, except those that are nonzero in the pruned model.

Another consistent finding is that sparse models tend to outperform dense ones for a fixed number of parameters. \citet{snip-followup} show that increasing the nominal size of ResNet-20 on CIFAR-10 while sparsifying to hold the number of parameters constant decreases the error rate. \citet{wavernn} obtain a similar result for audio synthesis, as do \citet{openai-block-sparse} for a variety of additional tasks across various domains. Perhaps most compelling of all are the many results, including in Figure~\ref{fig:arch_vs_prune}, showing that pruned models can obtain higher accuracies than the original models from which they are derived. This demonstrates that sparse models can not only outperform dense counterparts with the same number of parameters, but sometimes dense models with even more parameters.

\subsection{Pruning vs Architecture Changes}

One current unknown about pruning is how effective it tends to be relative to simply using a more efficient architecture. These options are not mutually exclusive, but it may be useful in guiding one's research or development efforts to know which choice is likely to have the larger impact. Along similar lines, it is unclear how pruned models from different architectures compare to one another---i.e., to what extent does pruning offer similar benefits across architectures?
To address these questions, we plotted the reported accuracies and compression/speedup levels of pruned models on ImageNet alongside the same metrics for different architectures with no pruning (Figure~\ref{fig:arch_vs_prune}).\footnote{
    Since many pruning papers report only change in accuracy or amount of pruning, without giving baseline numbers, we normalize all pruning results to have accuracies and model sizes/FLOPs as if they had begun with the same model. Concretely, this means multiplying the reported fraction of pruned size/FLOPs by a standardized initial value. This value is set to the median initial size or number of FLOPs reported for that architecture across all papers. This normalization scheme is not perfect, but does help control for different methods beginning with different baseline accuracies.
}
We plot results within a family of models as a single curve.\footnote{
    The EfficientNet family is given explicitly in the original paper \cite{efficientnet}, the ResNet family consists of ResNet-18, ResNet-34, ResNet-50, etc., and the VGG family consists of VGG-\{11, 13, 16, 19\}. There are no pruned EfficientNets since EfficientNet was published too recently. Results for non-pruned models are taken from \cite{efficientnet} and \cite{luigi}.
}


Figure~\ref{fig:arch_vs_prune} suggests several conclusions. First, it reinforces the conclusion that pruning can improve the time or space vs accuracy tradeoff of a given architecture, sometimes even increasing the accuracy. Second, it suggests that pruning generally does not help as much as switching to a better architecture. Finally, it suggests that pruning is more effective for architectures that are less efficient to begin with. 

%% file: is_problem.tex

While there do appear to be a few general and consistent findings in the pruning literature (see the previous section), by far the clearest takeaway is that pruning papers rarely make direct and controlled comparisons to existing methods. This lack of comparisons stems largely from a lack of experimental standardization and the resulting fragmentation in reported results. This fragmentation makes it difficult for even the most committed authors to compare to more than a few existing methods.


\subsection{Omission of Comparison}

Many papers claim to advance the state of the art, but don't compare to other methods---including many published ones---that make the same claim. 


\vspace{-2mm}
\paragraph{Ignoring Pre-2010s Methods}

There was already a rich body of work on neural network pruning by the mid 1990s (see, e.g., Reed's survey \cite{reed_pruning_1993}), which has been almost completely ignored except for Lecun's Optimal Brain Damage \cite{optimal-brain-damage} and Hassibi's Optimal Brain Surgeon \cite{optimal-brain-surgeon}. Indeed, multiple authors have rediscovered existing methods or aspects thereof, with \citet{learning-both} reintroducing the magnitude-based pruning of \citet{janowsky_pruning_1989}, \citet{snip} reintroducing the saliency heuristic of \citet{mozer_skeletonization:_1989}, and \citet{soft-filter-pruning} reintroducing the practice of ``reviving'' previously pruned weights described in \citet{early-brain-damage}.
\begin{figure}[t]
\begin{center}
\includegraphics[width=\linewidth]{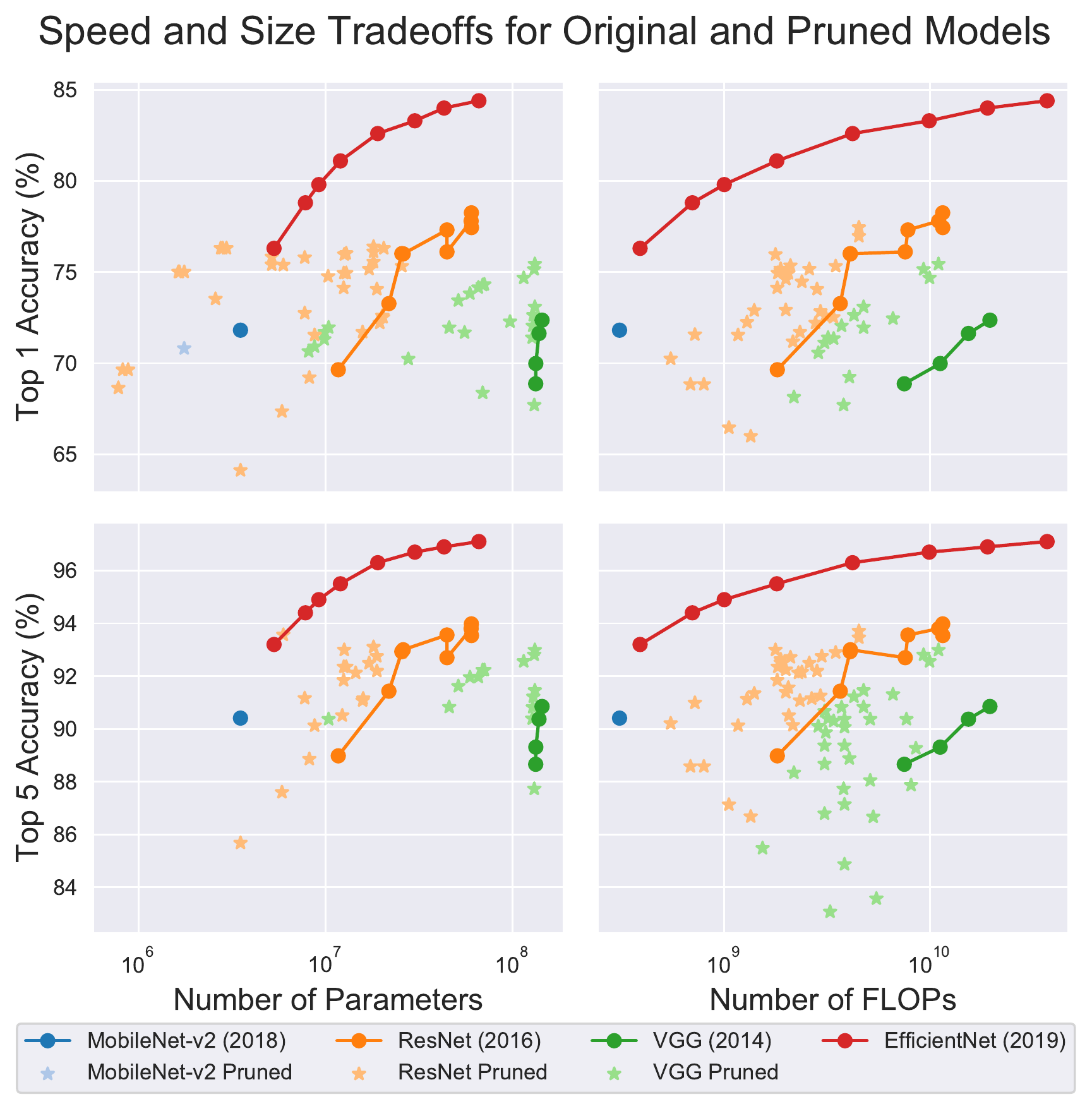}
\vspace{-3mm}
\caption{Size and speed vs accuracy tradeoffs for different pruning methods and families of architectures. Pruned models sometimes outperform the original architecture, but rarely outperform a better architecture.}
\label{fig:arch_vs_prune}
\end{center}
\end{figure}
\vspace{3mm}

\paragraph{Ignoring Recent Methods}

\begin{figure}[h]
\begin{center}
\includegraphics[width=\linewidth]{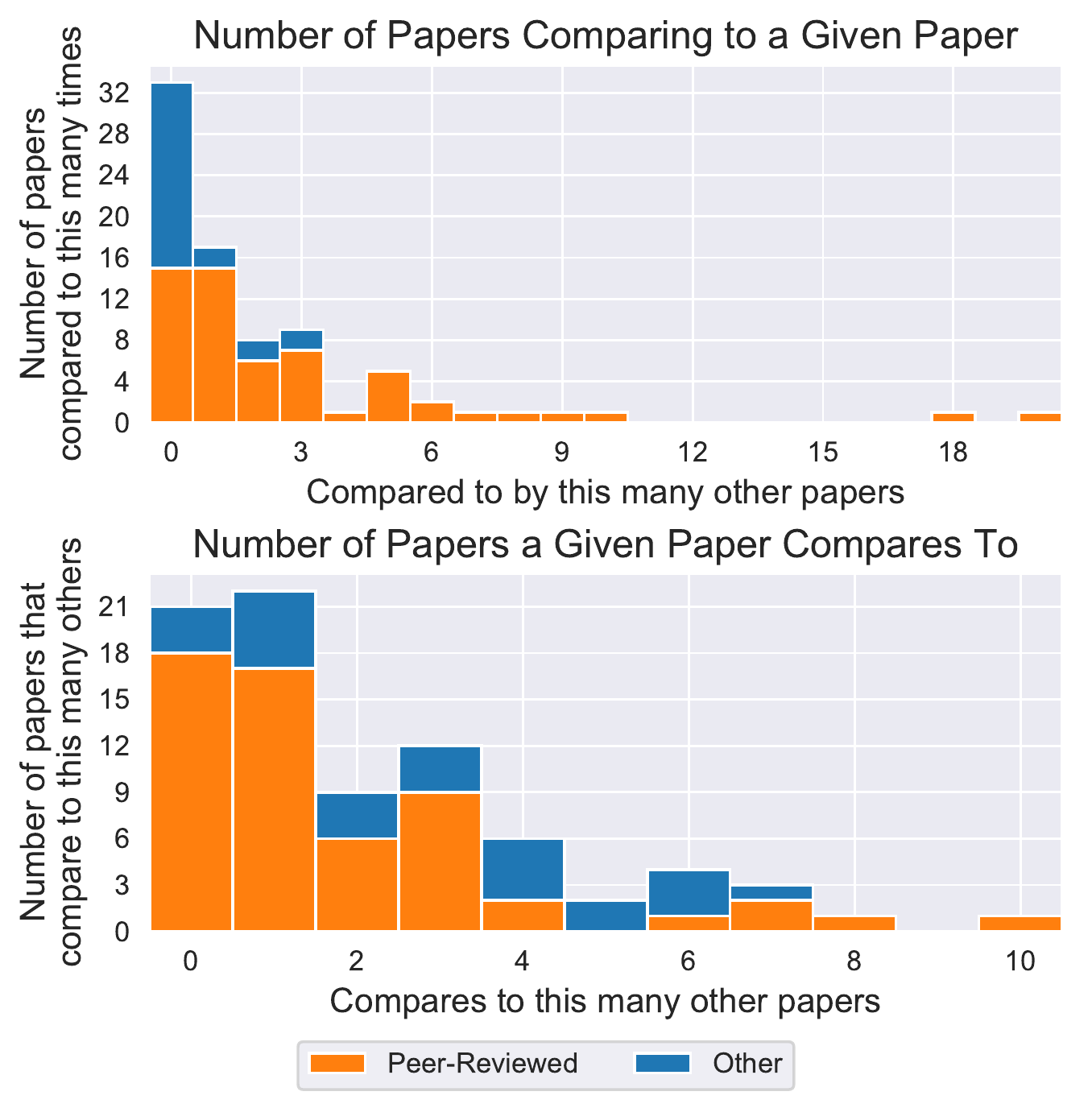}
\caption{Reported comparisons between papers.
}
\label{fig:paper_comparisons_hist}
\end{center}
\end{figure}

Even when considering only post-2010 approaches, there are still virtually no methods that have been shown to outperform all existing ``state-of-the-art'' methods. This follows from the fact, depicted in the top plot of Figure~\ref{fig:paper_comparisons_hist}, that there are dozens of modern papers---including many affirmed through peer review---that have never been compared to by any later study.

A related problem is that papers tend to compare to few existing methods. In the lower plot of Figure~\ref{fig:paper_comparisons_hist}, we see that more than a fourth of our corpus does not compare to any previously proposed pruning method, and another fourth compares to only one. Nearly all papers compare to three or fewer. This might be adequate if there were a clear progression of methods with one or two ``best'' methods at any given time, but this is not the case. 





\subsection{Dataset and Architecture Fragmentation}





Among \npapers papers, we found results using 49 datasets, 132 architectures, and 195 (dataset, architecture) combinations. As shown in Table~\ref{tbl:common_combos}, even the most common combination of dataset and architecture---VGG-16 on ImageNet\footnote{We adopt the common practice of referring to the ILSVRC2012 training and validation sets as ``ImageNet.''} \cite{imagenet}---is used in only 22 out of \npapers papers. Moreover, three of the top six most common combinations involve MNIST \cite{mnist}. As \citet{google-state-of-sparsity} and others have argued, using larger datasets and models is essential when assessing how well a method works for real-world networks. MNIST results may be particularly unlikely to generalize, since this dataset differs significantly from other popular datasets for image classification. In particular, its images are grayscale, composed mostly of zeros, and possible to classify with over 99\% accuracy using simple models \cite{mnist-page}.

\subsection{Metrics Fragmentation}

\begin{table}[t]
\begin{centering}
\begin{tabular}{ll|c}
\multicolumn{2}{c}{(Dataset, Architecture) Pair} & \shortstack{Number of Papers \\ using Pair} \\
\toprule
ImageNet & VGG-16 & 22 \\
ImageNet & ResNet-50 & 15 \\
MNIST & LeNet-5-Caffe & 14 \\
CIFAR-10 & ResNet-56 & 14 \\
MNIST & LeNet-300-100 & 12 \\
MNIST & LeNet-5 & 11 \\
ImageNet & CaffeNet & 10 \\
CIFAR-10 & CIFAR-VGG (Torch) & 8 \\
ImageNet & AlexNet & 8 \\
ImageNet & ResNet-18 & 6 \\
ImageNet & ResNet-34 & 6 \\
CIFAR-10 & ResNet-110 & 5 \\
CIFAR-10 & PreResNet-164 & 4 \\
CIFAR-10 & ResNet-32 & 4 \\
\end{tabular}
\end{centering}
\caption{All combinations of dataset and architecture used in at least 4 out of \npapers papers.
}
\label{tbl:common_combos}
\vspace*{1mm}
\end{table}

\begin{figure*}[hbt!]
\begin{center}
\includegraphics[width=\linewidth]{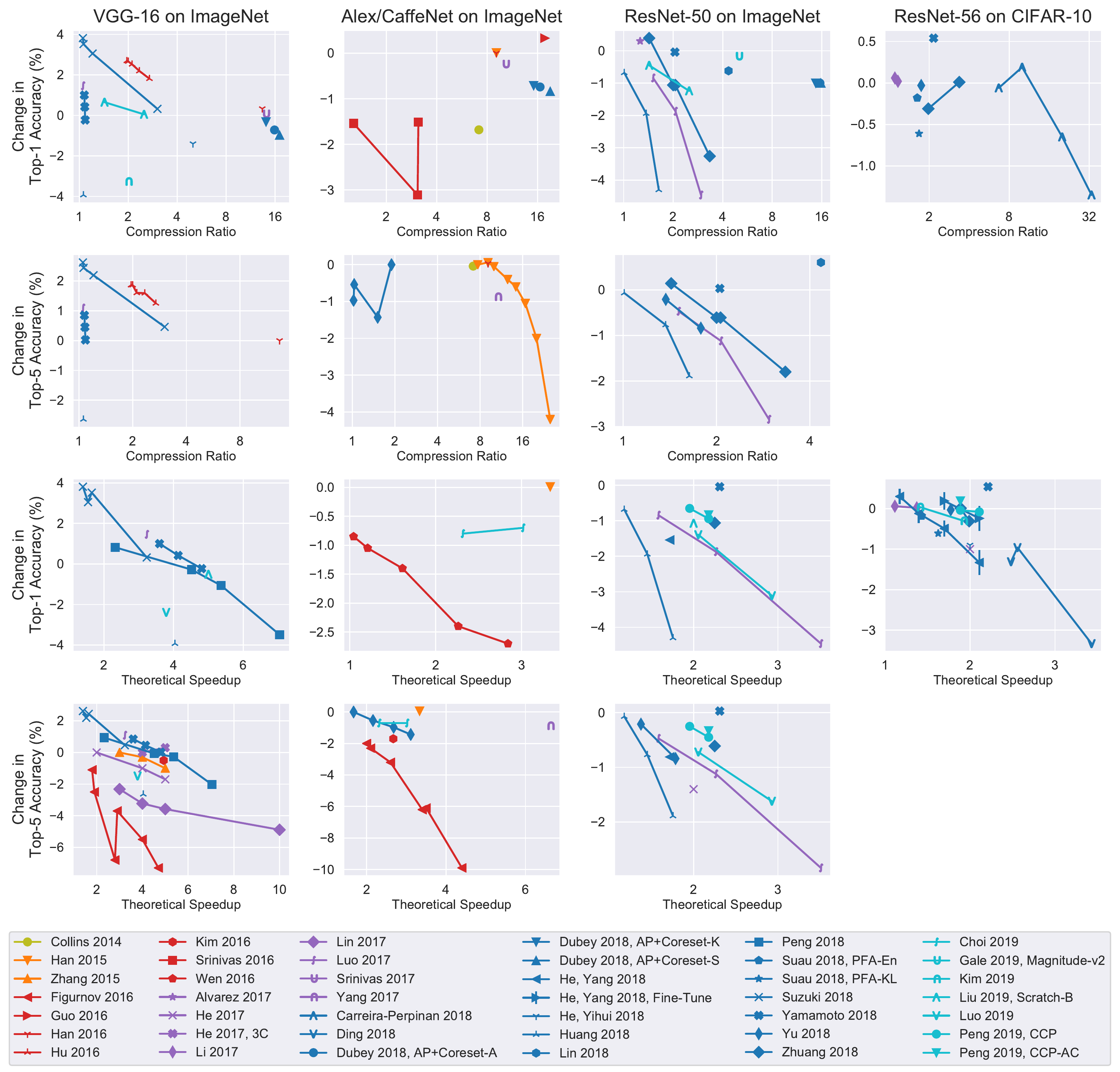}
\caption{Fragmentation of results. Shown are all self-reported results on the most common (dataset, architecture) combinations. Each column is one combination, each row shares an accuracy metric (y-axis), and pairs of rows share a compression metric (x-axis). Up and to the right is always better. Standard deviations are shown for He 2018 on CIFAR-10, which is the only result that provides any measure of central tendency.
As suggested by the legend, only 37 out of the \npapers papers in our corpus report any results using any of these configurations.}
\label{fig:prune_grid}
\end{center}
\vspace{-4mm}
\end{figure*}

As depicted in Figure~\ref{fig:prune_grid}, papers report a wide variety of metrics and operating points, making it difficult to compare results. Each column in this figure is one (dataset, architecture) combination taken from the four most common combinations\footnote{We combined the results for AlexNet and CaffeNet, which is a slightly modified version of AlexNet \cite{caffenet}, since many authors refer to the latter as ``AlexNet,'' and it is often unclear which model was used.}, excluding results on MNIST. Each row is one pair of metrics. Each curve is the efficiency vs accuracy tradeoff obtained by one method.\footnote{Since what counts as one method can be unclear, we consider all results from one paper to be one method except when two or more named methods within the paper report using at least one identical x-coordinate (i.e., when the paper's results can't be plotted as one curve).} Methods are color-coded by year.

It is hard to identify any consistent trends in these plots, aside from the existence of a tradeoff between efficiency and accuracy. A given method is only present in a small subset of plots. Methods from later years do not consistently outperform methods from earlier years. Methods within a plot are often incomparable because they report results at different points on the x-axis. Even when methods are nearby on the x-axis, it is not clear whether one meaningfully outperforms another since neither reports a standard deviation or other measure of central tendency. Finally, most papers in our corpus do not report any results with any of these common configurations.

\subsection{Incomplete Characterization of Results}

\begin{figure}[h]
\begin{center}
\includegraphics[width=\linewidth]{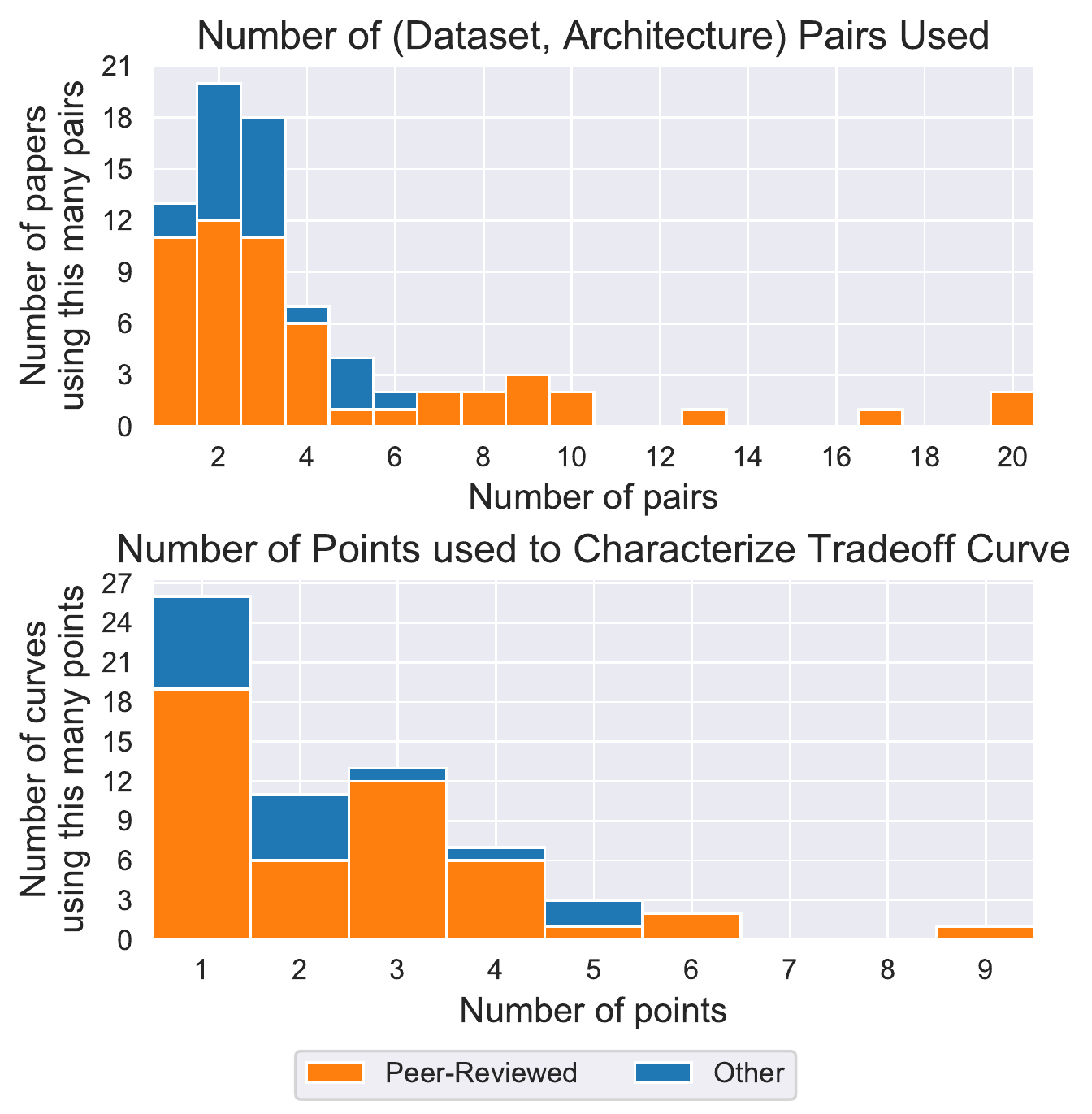}
\caption{Number of results reported by each paper, excluding MNIST. \textit{Top)} Most papers report on three or fewer (dataset, architecture) pairs. \textit{Bottom)} For each pair used, most papers characterize their tradeoff between amount of pruning and accuracy using a single point in the efficiency vs accuracy curve. In both plots, the pattern holds even for peer-reviewed papers.}
\label{fig:numresults_stats}
\end{center}
\end{figure}

If all papers reported a wide range of points in their tradeoff curves across a large set of models and datasets, there might be some number of direct comparisons possible between any given pair of methods. 
As we see in the upper half of Figure~\ref{fig:numresults_stats}, however, most papers use at most three (dataset, architecture) pairs; and as we see in the lower half, they use at most three---and often just one---point to characterize each curve. Combined with the fragmentation in experimental choices, this means that different methods' results are rarely directly comparable. Note that the lower half restricts results to the four most common (dataset, architecture) pairs.

\subsection{Confounding Variables} \label{sec:confounding}

Even when comparisons include the same datasets, models, metrics, and operating points, other confounding variables still make meaningful comparisons difficult. Some variables of particular interest include:
\begin{itemize}[leftmargin=4mm]
    \itemsep2pt
    \vspace{-3.5mm}
    \item Accuracy and efficiency of the initial model
    \item Data augmentation and preprocessing
    \item Random variations in initialization, training, and fine-tuning. This includes choice of optimizer, hyperparameters, and learning rate schedule.
    \item Pruning and fine-tuning schedule
    \item Deep learning library. Different libraries are known to yield different accuracies for the same architecture and dataset \cite{unreproducibleCurtis, unreproducible3} and may have subtly different behaviors \cite{kerasBnWeird}.
    \item Subtle differences in code and environment that may not be easily attributable to any of the above variations \cite{unreproducible0, unreproducible1, unreproducible4}.
\end{itemize}

In general, it is not clear that any paper can succeed in accounting for all of these confounders unless that paper has both used the same code as the methods to which it compares and reports enough measurements to average out random variations. This is exceptionally rare, with \citet{google-state-of-sparsity} and \citet{rethinking-net-pruning} being arguably the only examples. Moreover, neither of these papers introduce novel pruning methods \textit{per se} but are instead inquiries into the efficacy of existing methods.


Many papers attempt to account for subsets of these confounding variables. A near universal practice in this regard is reporting change in accuracy relative to the original model, in addition to or instead of raw accuracy. This helps to control for the accuracy of the initial model. However, as we demonstrate in Section~\ref{sec:bench}, this is not sufficient to remove initial model as a confounder. Certain initial models can be pruned more or less efficiently, in terms of the accuracy vs compression tradeoff. This holds true even with identical pruning methods and all other variables held constant.




\vspace{2mm}
There are at least two more empirical reasons to believe that confounding variables can have a significant impact. First, as one can observe in Figure~\ref{fig:prune_grid}, methods often introduce changes in accuracy of much less than 1\% at reported operating points. This means that, even if confounders have only a tiny impact on accuracy, they can still have a large impact on which method appears better. 

\begin{figure}[h]
\begin{center}
\includegraphics[width=\linewidth]{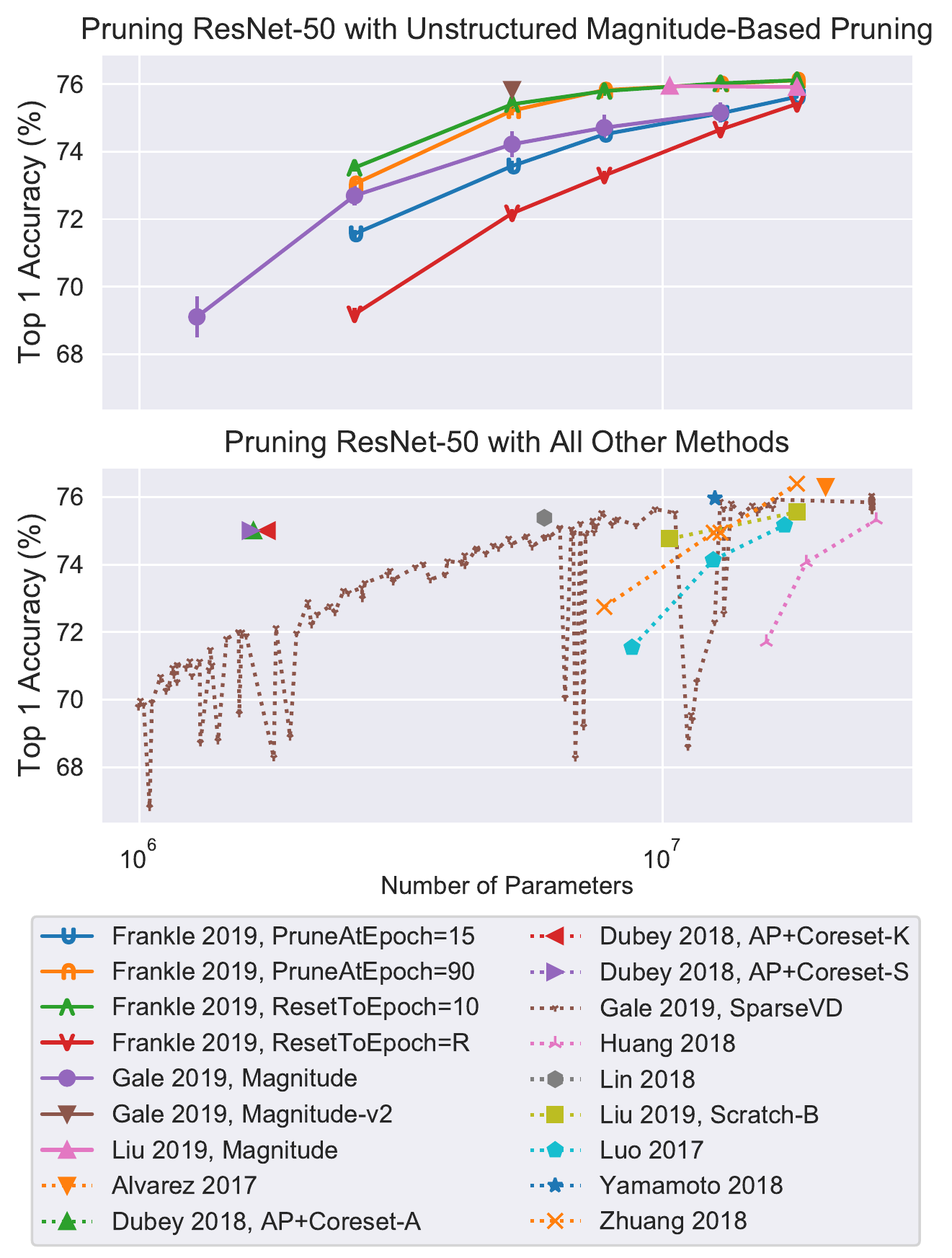}
\caption{Pruning ResNet-50 on ImageNet. Methods in the upper plot all prune weights with the smallest magnitudes, but differ in implementation, pruning schedule, and fine-tuning. The variation caused by these variables is similar to the variation across different pruning methods, whose results are shown in the lower plot. All results are taken from the original papers.}
\label{fig:magnitude_vs_nonmagnitude}
\end{center}
\vspace*{3mm}
\end{figure}

\vspace{2mm}
Second, as shown in Figure~\ref{fig:magnitude_vs_nonmagnitude}, existing results demonstrate that different training and fine-tuning settings can yield nearly as much variability as different methods. Specifically, consider 1) the variability introduced by different fine-tuning methods for unstructured magnitude-based pruning (Figure 6 top) and 2) the variability introduced by entirely different pruning methods (Figure 6 bottom). The variability between fine-tuning methods is nearly as large as the variability between pruning methods.

%% file: why_problem.tex
In the previous section, we discussed the fragmentation of datasets, models, metrics, operating points, and experimental details, and how this fragmentation makes evaluating the efficacy of individual pruning methods difficult. In this section, we argue that there are additional barriers to comparing methods that stem from common practices in how methods and results are presented.

\subsection{Architecture Ambiguity}

It is often difficult, or even impossible, to identify the exact architecture that authors used. Perhaps the most prevalent example of this is when authors report using some sort of ResNet \cite{resnet, resnet2}. Because there are two different variations of ResNets, introduced in these two papers, saying that one used a ``ResNet-50'' is insufficient to identify a particular architecture. Some authors do appear to deliberately point out the type of ResNet they use (e.g., \cite{network-slimming, more-is-less}). However, given that few papers even hint at the possibility of confusion, it seems unlikely that all authors are even aware of the ambiguity, let alone that they have cited the corresponding paper in all cases. 


Perhaps the greatest confusion is over VGG networks \cite{vgg}. Many papers describe experimenting on ``VGG-16,'' ``VGG,'' or ``VGGNet,'' suggesting a standard and well-known architecture. In many cases, what is actually used is a custom variation of some VGG model, with removed fully-connected layers \cite{google-interchannel, thinet-channel-norms}, smaller fully-connected layers \cite{snip}, or added dropout or batchnorm \cite{network-slimming, snip, extreme-net-compress,sparse-variational-dropout, ding-auto-balanced, apple-pfa}.

In some cases, papers simply fail to make clear what model they used (even for non-VGG architectures). For example, one paper just states that their segmentation model ``is composed from an inception-like network branch and a DenseNet network branch.'' Another paper attributes their VGGNet to \cite{deepFaceRecognition}, which mentions three VGG networks. \citet{rethinking-net-pruning} and \citet{lottery-ticket} have circular references to one another that can no longer be resolved because of simultaneous revisions. One paper mentions using a ``VGG-S'' from the Caffe Model Zoo, but as of this writing, no model with this name exists there. Perhaps the most confusing case is the Lenet-5-Caffe reported in one 2017 paper. The authors are to be commended for explicitly stating not only that they use Lenet-5-Caffe, but their exact architecture. However, they describe an architecture with an 800-unit fully-connected layer, while examination of both the Caffe \texttt{.prototxt} files \cite{lenet-5-proto1, lenet-5-proto2} and associated blog post \cite{lenet-5-caffe} indicates that no such layer exists in Lenet-5-Caffe.


\subsection{Metrics Ambiguity} \label{sec:otherAmbiguity}


It can also be difficult to know what the reported metrics mean. For example, many papers include a metric along the lines of ``Pruned\%''. In some cases, this means fraction of the parameters or FLOPs remaining \cite{apple-pfa}. In other cases, it means the fraction of parameters or FLOPs removed \cite{learning-both, lempitsky-fast-convnets, balanced-sparsity}. There is also widespread misuse of the term ``compression ratio,'' which the compression literature has long used to mean $\frac{\text{original size}}{\text{compressed size}}$ \cite{bbp, pfor, groupSimd, zfp, dfcm, sprintz}, but many pruning authors define (usually without making the formula explicit) as $1 - \frac{\text{compressed size}}{\text{original size}}$.

Reported ``speedup'' values present similar challenges. These values are sometimes wall time, sometimes original number of FLOPs divided by pruned number of FLOPs, sometimes a more complex formula relating these two quantities \cite{more-is-less, soft-filter-pruning}, and sometimes never made clear. Even when reporting FLOPs, which is nominally a consistent metric, different authors measure it differently (e.g., \cite{nvidia-taylor-pruning} vs \cite{convnet-tensor-decomp}), though most often papers entirely omit their formula for computing FLOPs. We found up to a factor of four variation in the reported FLOPs of different papers for the same architecture and dataset, with \cite{sze-energy-aware} reporting 371 MFLOPs for AlexNet on ImageNet, \cite{samsung-winograd-sparse} reporting 724 MFLOPs, and \cite{learning-both} reporting 1500 MFLOPs.

%% file: recommendations.tex
In the previous sections, we have argued that existing work tends to
\begin{itemize}[leftmargin=4mm]
    \itemsep0pt
    \vspace{-3mm}
    \item make it difficult to identify the exact experimental setup and metrics,
    \item use too few (dataset, architecture) combinations,
    \item report too few points in the tradeoff curve for any given combination, and no measures of central tendency,
    \item omit comparison to many methods that might be state-of-the-art, and
    \item fail to control for confounding variables.
    \vspace{-3mm}
\end{itemize}
These problems often make it difficult or impossible to assess the relative efficacy of different pruning methods.
To enable direct comparison between methods in the future, we suggest the following practices:
\begin{itemize}[leftmargin=4mm]
    \itemsep-.5pt
    \vspace{-2mm}
\item Identify the \textit{exact} sets of architectures, datasets, and metrics used, ideally in a structured way that is not scattered throughout the results section. 
\item Use at least three (dataset, architecture) pairs, including modern, large-scale ones. MNIST and toy models do not count. AlexNet, CaffeNet, and Lenet-5 are no longer modern architectures.
\item For any given pruned model, report both compression ratio and theoretical speedup. Compression ratio is defined as the original size divided by the new size. Theoretical speedup is defined as the original number of multiply-adds divided by the new number. Note that there is no reason to report only one of these metrics.
\item For ImageNet and other many-class datasets, report both Top-1 and Top-5 accuracy. There is again no reason to report only one of these.
\item Whatever metrics one reports for a given pruned model, also report these metrics for an appropriate control (usually the original model before pruning).
\item Plot the tradeoff curve for a given dataset and architecture, alongside the curves for competing methods. 
\item When plotting tradeoff curves, use at least 5 operating points spanning a range of compression ratios. The set of ratios $\{2, 4, 8, 16, 32\}$ is a good choice.
\item Report and plot means and sample standard deviations, instead of one-off measurements, whenever feasible.
\item Ensure that all methods being compared use identical libraries, data loading, and other code to the greatest extent possible.
\vspace{-2mm}
\end{itemize}

We also recommend that reviewers demand a much greater level of rigor when evaluating papers that claim to offer a better method of pruning neural networks.

%% file: shrinkbench_v2.tex
\newcommand{\SB}{ShrinkBench}
\newcommand{\NEXP}{800} 

\vspace{1mm}



\subsection{Overview of ShrinkBench}

To make it as easy as possible for researchers to put our suggestions into practice, we have created an open-source library for pruning called ShrinkBench. 
ShrinkBench provides standardized and extensible functionality for training, pruning, fine-tuning, computing metrics, and plotting, all using a standardized set of pretrained models and datasets.

ShrinkBench is based on PyTorch \cite{pytorch} and is designed to allow easy evaluation of methods with arbitrary scoring functions, allocation of pruning across layers, and sparsity structures. In particular, given a callback defining how to compute masks for a model's parameter tensors at a given iteration, ShrinkBench will automatically apply the pruning, update the network according to a standard training or fine-tuning setup, and compute metrics across many models, datasets, random seeds, and levels of pruning.
We defer discussion of ShrinkBench's implementation and API to the project's documentation.

\subsection{Baselines}

We used ShrinkBench to implement several existing pruning heuristics, both as examples of how to use our library and as baselines that new methods can compare to:
\begin{itemize}[leftmargin=4mm]
    \itemsep-1pt
    \vspace{-2mm}
    \item \textbf{Global Magnitude Pruning} - prunes the weights with the lowest absolute value anywhere in the network.
    \item \textbf{Layerwise Magnitude Pruning} - for each layer, prunes the weights with the lowest absolute value.
    \item \textbf{Global Gradient Magnitude Pruning} - prunes the weights with the lowest absolute value of (weight $\times$ gradient), evaluated on a batch of inputs.
    \item \textbf{Layerwise Gradient Magnitude Pruning} - for each layer, prunes the weights the lowest absolute value of (weight $\times$ gradient), evaluated on a batch of inputs.
    \item \textbf{Random Pruning} - prunes each weight independently with probability equal to the fraction of the network to be pruned.
\end{itemize}
\vspace{-2mm}

Magnitude-based approaches are common baselines in the literature and have been shown to be competitive with more complex methods \cite{learning-both, han-prune-quant-huff, google-state-of-sparsity, lottery-ticket-followup}. Gradient-based methods are less common, but are simple to implement and have recently gained popularity \cite{snip, snip-followup, nisp}. Random pruning is a common straw man that can serve as a useful debugging tool. Note that these baselines are not reproductions of any of these methods, but merely inspired by their pruning heuristics.

\subsection{Avoiding Pruning Pitfalls with Shrinkbench}

Using the described baselines, we pruned over \NEXP{} networks with varying datasets, networks, compression ratios, initial weights and random seeds.
In doing so, we identified various pitfalls associated with experimental practices that are currently common in the literature but are avoided by using ShrinkBench.

We highlight several noteworthy results below. For additional experimental results and details, see Appendix~\ref{apx:res}.  One standard deviation bars across three runs are shown for all CIFAR-10 results.

\begin{figure}[h]
\begin{center}
\includegraphics[width=\linewidth]{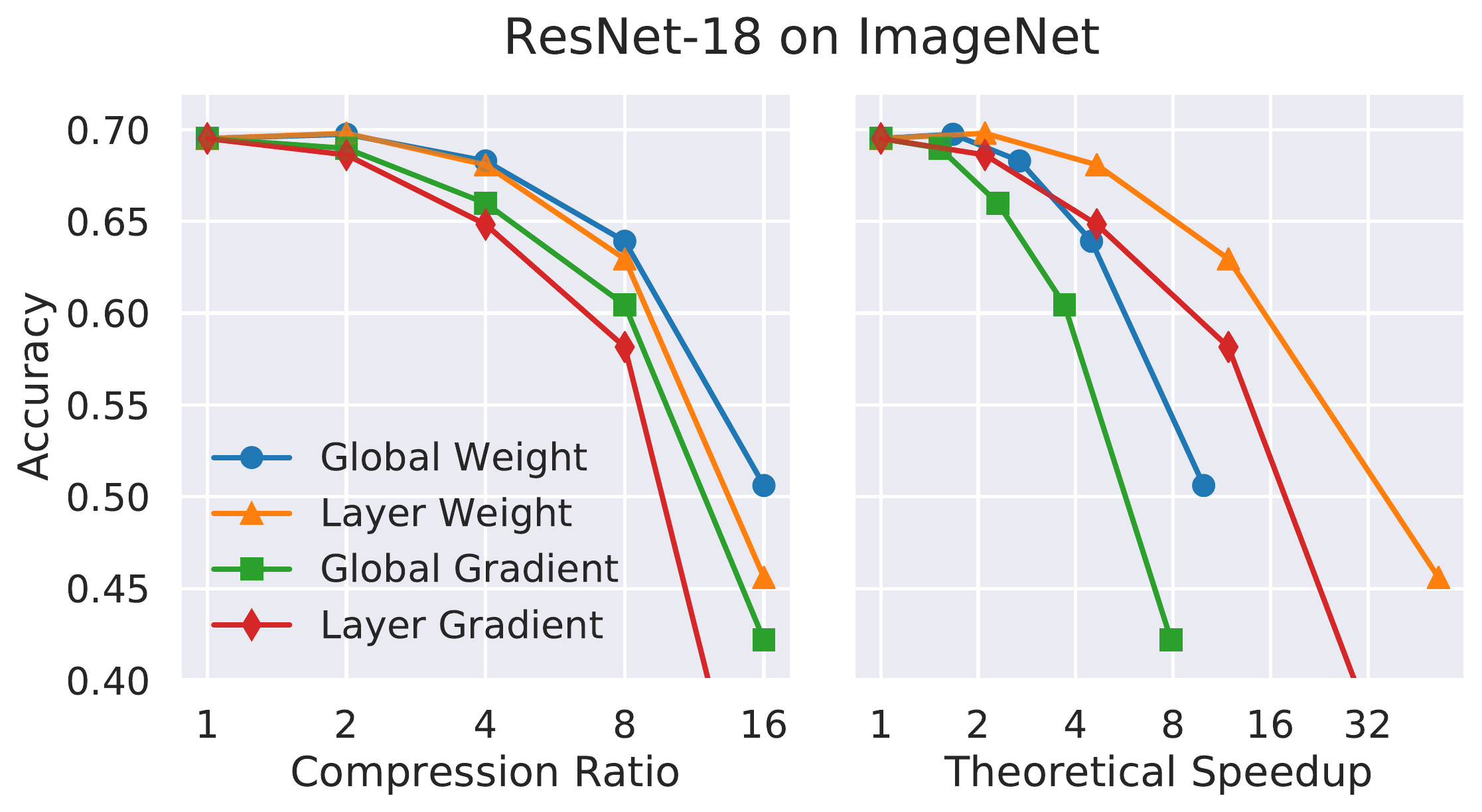}
\caption{Top 1 Accuracy for ResNet-18 on ImageNet for several compression ratios and their corresponding theoretical speedups.
Global methods give higher accuracy than Layerwise ones for a fixed model size, but the reverse is true for a fixed theoretical speedup.
}
\label{fig:ImageNet_flops}
\end{center}
\vspace*{-3mm}
\end{figure}

\vspace{-2mm}
\paragraph{Metrics are not Interchangeable.}
As discussed previously, it is common practice to report either reduction in the number of parameters or in the number of FLOPs. If these metrics are extremely correlated, reporting only one is sufficient to characterize the efficacy of a pruning method. We found after computing these metrics for the same model under many different settings that reporting one metric is not sufficient. While these metrics are correlated, the correlation is different for each pruning method. Thus, the relative performance of different methods can vary significantly under different metrics (Figure~\ref{fig:ImageNet_flops}).

\vspace{-3mm}
\paragraph{Results Vary Across Models, Datasets, and Pruning Amounts}

Many methods report results on only a small number of datasets, models, amounts of pruning, and random seeds. If the relative performance of different methods tends to be constant across all of these variables, this may not be problematic. However, our results suggest that this performance is not constant.

\begin{figure}[h]
\begin{center}
\includegraphics[width=\linewidth]{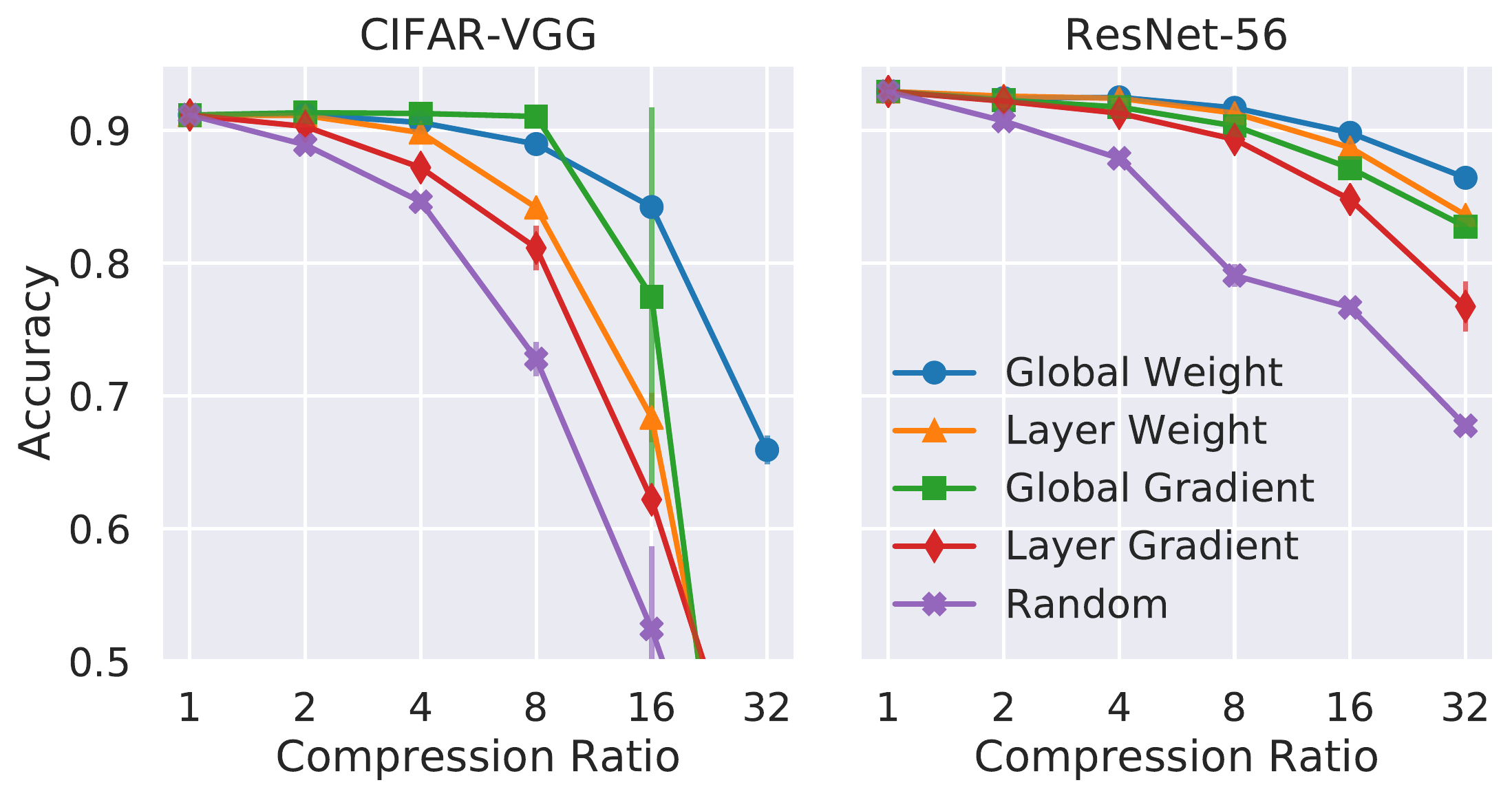}
\caption{Top 1 Accuracy on CIFAR-10 for several compression ratios.
 Global Gradient performs better than Global Magnitude for CIFAR-VGG on low compression ratios, but worse otherwise.
 Global Gradient is consistently better than Layerwise Magnitude on CIFAR-VGG, but consistently worse on ResNet-56.
 %
}
\label{fig:sb_CIFAR10}
\end{center}
\end{figure}

Figure \ref{fig:sb_CIFAR10} shows the accuracy for various compression ratios for CIFAR-VGG \cite{vggCifarTorch} and ResNet-56 on CIFAR-10.
In general, Global methods are more accurate than Layerwise methods and Magnitude-based methods are more accurate than Gradient-based methods, with random performing worst of all.
However, if one were to look only at CIFAR-VGG for compression ratios smaller than 10, one could conclude that Global Gradient outperforms all other methods.
Similarly, while Global Gradient consistently outperforms Layerwise Magnitude on CIFAR-VGG, the opposite holds on ResNet-56 (i.e., the orange and green lines switch places).
%
%
%

Moreover, we found that for some settings close to the drop-off point (such as Global Gradient, compression 16), different random seeds yielded significantly different results (0.88 vs 0.61 accuracy) due to the randomness in minibatch selection. This is illustrated by the large vertical error bar in the left subplot.

%
%

\vspace{-4mm}
\paragraph{Using the Same Initial Model is Essential.}

As mentioned in Section~\ref{sec:confounding}, many methods are evaluated using different initial models with the same architecture. To assess whether beginning with a different model can skew the results, we created two different models and evaluated Global vs Layerwise Magnitude pruning on each with all other variables held constant.

\begin{figure}[t]
\begin{center}
\includegraphics[width=\linewidth]{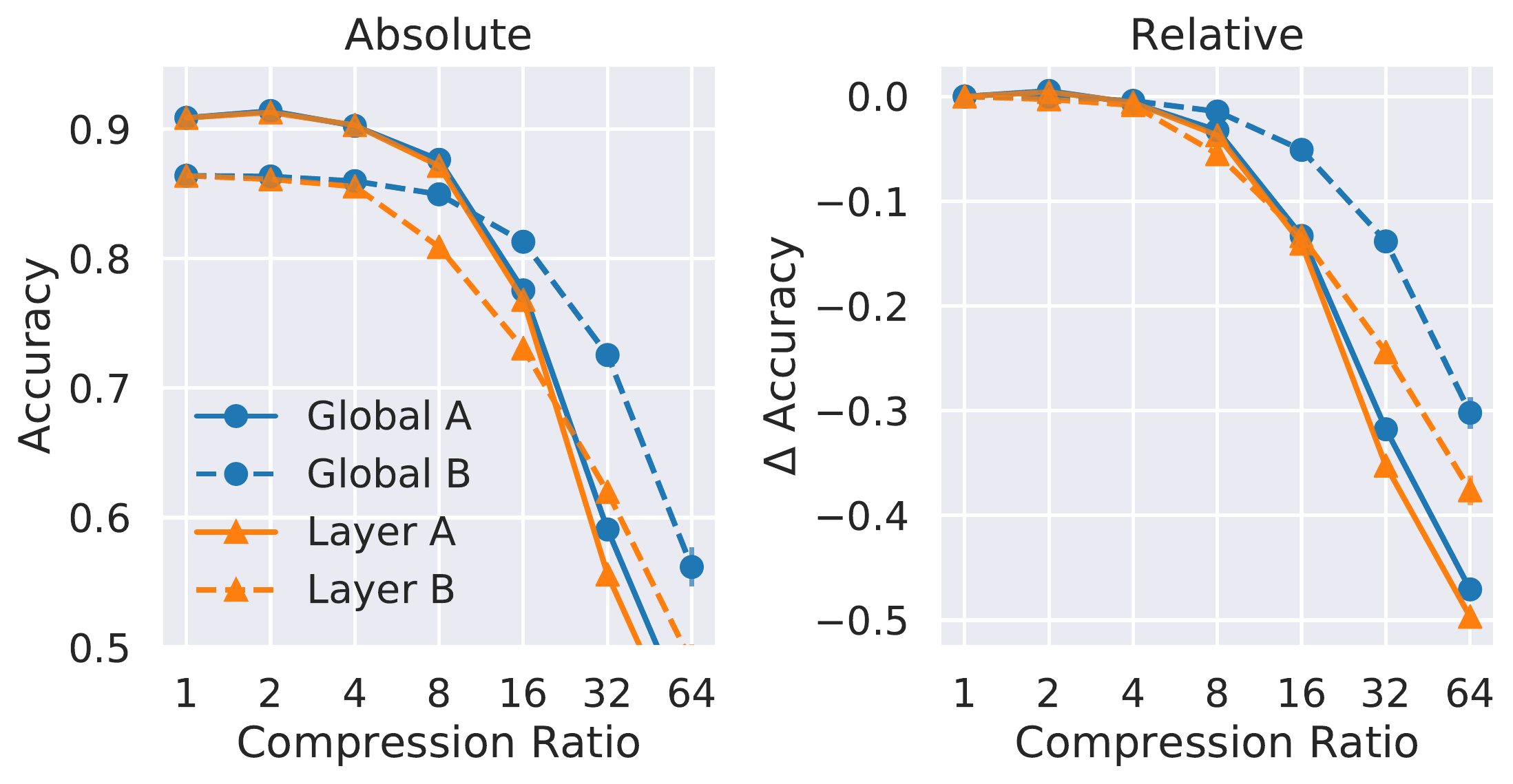}
\caption{Global and Layerwise Magnitude Pruning on two different ResNet-56 models.
%
%
Even with all other variables held constant, different initial models yield different tradeoff curves. This may cause one method to erroneously appear better than another. Controlling for initial accuracy does not fix this. 
}
\label{fig:sb_resnet56_confounding}
\end{center}
\end{figure}

To obtain the models, we trained two ResNet-56 networks using Adam until convergence with $\eta=10^{-3}$ and $\eta=10^{-4}$.
We'll refer to these pretrained weights as Weights A and Weights B, respectively.
%
As shown on the left side of Figure~\ref{fig:sb_resnet56_confounding}, the different methods appear better on different models. With Weights A, the methods yield similar absolute accuracies. With Weights B, however, the Global method is more accurate at higher compression ratios.

%
%
%

We also found that the common practice of examining changes in accuracy is insufficient to correct for initial model as a confounder. Even when reporting changes, one pruning method can artificially appear better than another by virtue of beginning with a different model. We see this on the right side of Figure~\ref{fig:sb_resnet56_confounding}, where Layerwise Magnitude with Weights B appears to outperform Global Magnitude with Weights A, even though the former never outperforms the latter when initial model is held constant.



%% file: appendix_corpus.tex
\section{Corpus and Data Cleaning} \label{sec:corpus}


We selected the \npapers papers used in our analysis in the following way. First, we conducted an ad hoc literature search, finding widely cited papers introducing pruning methods and identifying other pruning papers that cited them using Google Scholar. We then went through the conference proceedings from the past year's NeurIPS, ICML, CVPR, ECCV, and ICLR and added all relevant papers (though it is possible we had false dismissals if the title and abstract did not seem relevant to pruning). Finally, during the course of cataloging which papers compared to which others, we added to our corpus any pruning paper that at least one existing paper in our corpus purported to compare to. We included both published papers and unpublished ones of reasonable quality (typically on arXiv). Since we make strong claims about the lack of comparisons, we included in our corpus five papers whose methods technically do not meet our definition of pruning but are similar in spirit and compared to by various pruning papers.
In short, we included essentially every paper introducing a method of pruning neural networks that we could find, taking care to capture the full directed graph of papers and comparisons between them.







Because different papers report slightly different metrics, particularly with respect to model size, we converted reported results to a standard set of metrics whenever possible. For example, we converted reported Top-1 error rates to Top-1 accuracies, and fractions of parameters pruned to compression ratios. Note that it is not possible to convert between size metrics and speedup metrics, since the amount of computation associated with a given parameter can depend on the layer in which it resides (since convolutional filters are reused at many spatial positions). For simplicity and uniformity, we only consider self-reported results except where stated otherwise.

We also did not attempt to capture all reported metrics, but instead focused only on model size reduction and theoretical speedup, since 1) these are by far the most commonly reported and, 2) there is already a dearth of directly comparable numbers even for these common metrics. This is not entirely fair to methods designed to optimize other metrics, such as power consumption \cite{bayesian-compression, sze-energy-aware, learning-both, samsung-vbmf-tucker}, memory bandwidth usage \cite{extreme-net-compress, samsung-vbmf-tucker}, or fine-tuning time \cite{uiuc-coreset-pruning, pcas, sss, soft-filter-pruning}, and we consider this a limitation of our analysis.

Lastly, as a result of relying on reading of hundreds of pages of dense technical content, we are confident that we have made some number of isolated errors. We therefore welcome correction by email and refer the reader to the arXiv version of this paper for the most up-to-date revision.

%% file: appendix_checklist.tex
\section{Checklist for Evaluating a Pruning Method} \label{apx:recommendations}

\noindent For any pruning technique proposed, check if:
\begin{itemize}
\item It is contextualized with respect to magnitude pruning, recently-published pruning techniques, and pruning techniques proposed prior to the 2010s.
\item The pruning algorithm, constituent subroutines (e.g., score, pruning, and fine-tuning functions), and hyperparameters are presented in enough detail for a reader to reimplement and match the results in the paper.
\item All claims about the technique are appropriately restricted to only the experiments presented (e.g., CIFAR-10, ResNets, image classification tasks, etc.).
\item There is a link to downloadable source code.
\end{itemize}

\noindent For all experiments, check if you include:
\begin{itemize}
\item A detailed description of the architecture with hyperparameters in enough detail to for a reader to reimplement it and train it to the same performance reported in the paper.
\item If the architecture is not novel: a citation for the architecture/hyperparameters and a description of any differences in architecture, hyperparameters, or performance in this paper.
\item A detailed description of the dataset hyperparameters (e.g., batch size and augmentation regime) in enough detail for a reader to reimplement it.
\item A description of the library and hardware used.
\end{itemize}

\noindent For all results, check if:
\begin{itemize}
\item Data is presented across a range of compression ratios, including extreme compression ratios at which the accuracy of the pruned network declines substantially.
\item Data specifies the raw accuracy of the network at each point.
\item Data includes multiple runs with separate initializations and random seeds.
\item Data includes clearly defined error bars and a measure of central tendency (e.g., mean) and variation (e.g., standard deviation).
\item Data includes FLOP-counts if the paper makes arguments about efficiency and performance due to pruning.
\end{itemize}

\noindent For all pruning results presented, check if there is a comparison to:
\begin{itemize}
\item A random pruning baseline.
    \begin{itemize}
    \item A global random pruning baseline.
    \item A random pruning baseline with the same layerwise pruning proportions as the proposed technique.
    \end{itemize}
\item A magnitude pruning baseline.
    \begin{itemize}
    \item A global or uniform layerwise proportion magnitude pruning baseline.
    \item A magnitude pruning baseline with the same layerwise pruning proportions as the proposed technique.
    \end{itemize}
\item Other relevant state-of-the-art techniques, including:
    \begin{itemize}
    \item A description of how the comparisons were produced (data taken from paper, reimplementation, or reuse of code from the paper) and any differences or uncertainties between this setting and the setting used in the main experiments.
    \end{itemize}
\end{itemize}

%% file: appendix_framework.tex

\section{Experimental Setup} \label{apx:exp}

For reproducibility purposes, \SB{} fixes random seeds for all the dependencies (\texttt{PyTorch}, \texttt{NumPy}, \texttt{Python}).

\subsection{Pruning Methods}

For the reported experiments, we did not prune the classifier layer preceding the softmax.
\SB{} supports pruning said layer as an option to all proposed pruning strategies.
For both Global and Layerwise Gradient Magnitude Pruning a single minibatch is used to compute the gradients for the pruning.
Three independent runs using different random seeds were performed for every CIFAR10 experiment.
We found some variance across methods that relied on randomness, such as random pruning or gradient based methods that use a sampled minibatch to compute the gradients with respect to the weights.

\subsection{Finetuning Setup}

Pruning was performed from the pretrained weights and fixed from there forwards.
%
%
Early stopping is implemented during finetuning.
Thus if the validation accuracy repeatedly decreases after some point we stop the finetuning process to prevent overfitting.
%

All reported CIFAR10 experiments used the following finetuning setup:
\begin{itemize}[leftmargin=4mm]
    \itemsep-2pt
    \vspace{-2mm}
    \item Batch size: 64
    \item Epochs: 30
    \item Optimizer: Adam
    \item Initial Learning Rate: $3 \times 10^{-4}$
    \item Learning rate schedule: Fixed
\end{itemize}
All reported ImageNet experiments used the following finetuning setup
\begin{itemize}[leftmargin=4mm]
    \itemsep-2pt
    \vspace{-2mm}
    \item Batch size: 256
    \item Epochs: 20
    \item Optimizer: SGD with Nesterov Momentum (0.9)
    \item Initial Learning Rate: $1 \times 10^{-3}$
    \item Learning rate schedule: Fixed
\end{itemize}

%% file: appendix_results.tex

\section{Additional Results} \label{apx:res}

Here we include the entire set of results obtained with \SB.
For CIFAR10, results are included for CIFAR-VGG, ResNet-20, ResNet-56 and ResNet-110. Standard deviations across three different random runs are plotted as error bars.
For ImageNet, results are reported for ResNet-18.

\clearpage

\begin{figure*}
\begin{minipage}[b]{.45\textwidth}
\centering
\includegraphics[width=\linewidth]{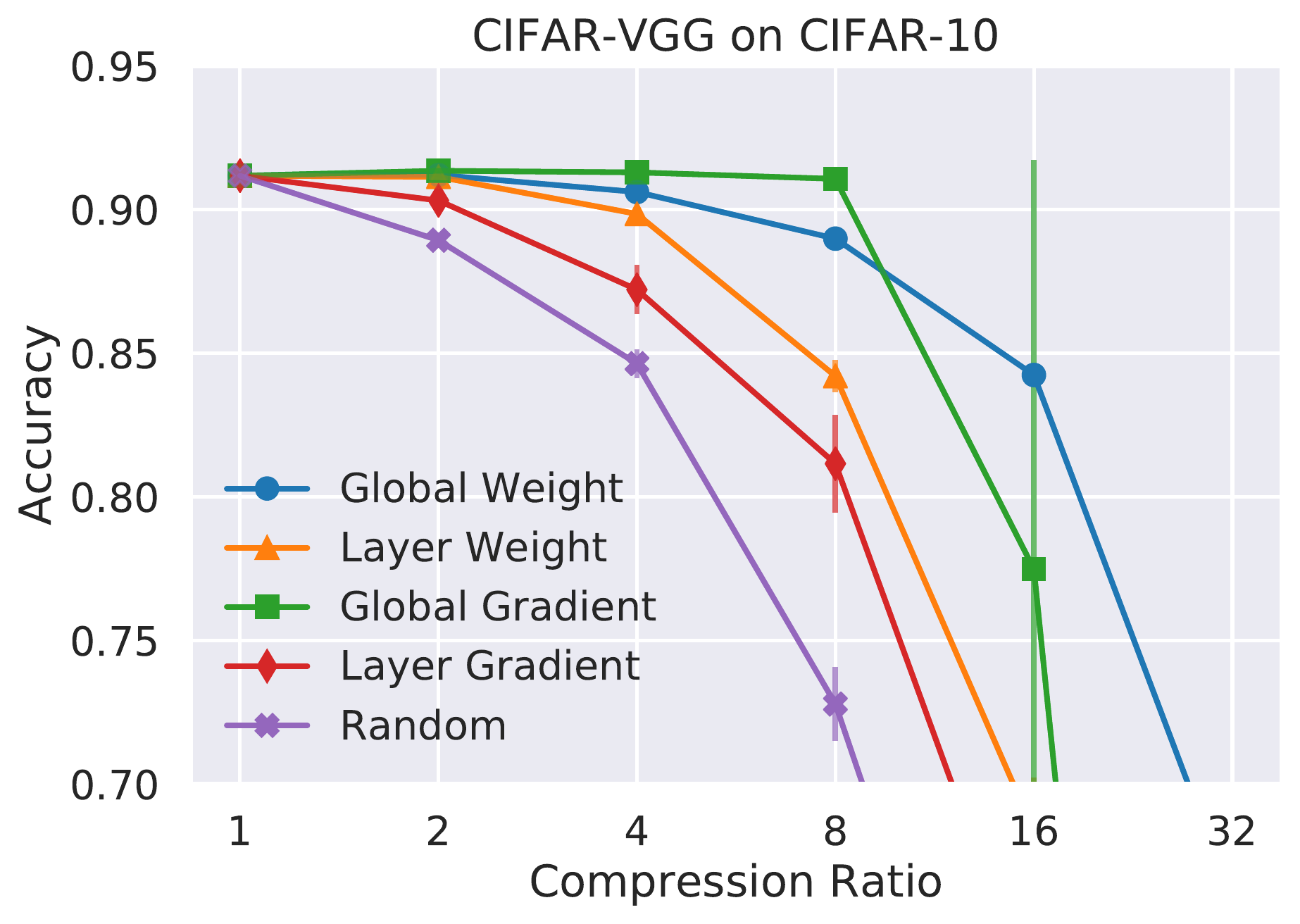}
\captionof{figure}{Accuracy for several levels of compression for CIFAR-VGG on CIFAR-10}
\end{minipage}
\hfill
\begin{minipage}[b]{.45\textwidth}
\centering
\includegraphics[width=\linewidth]{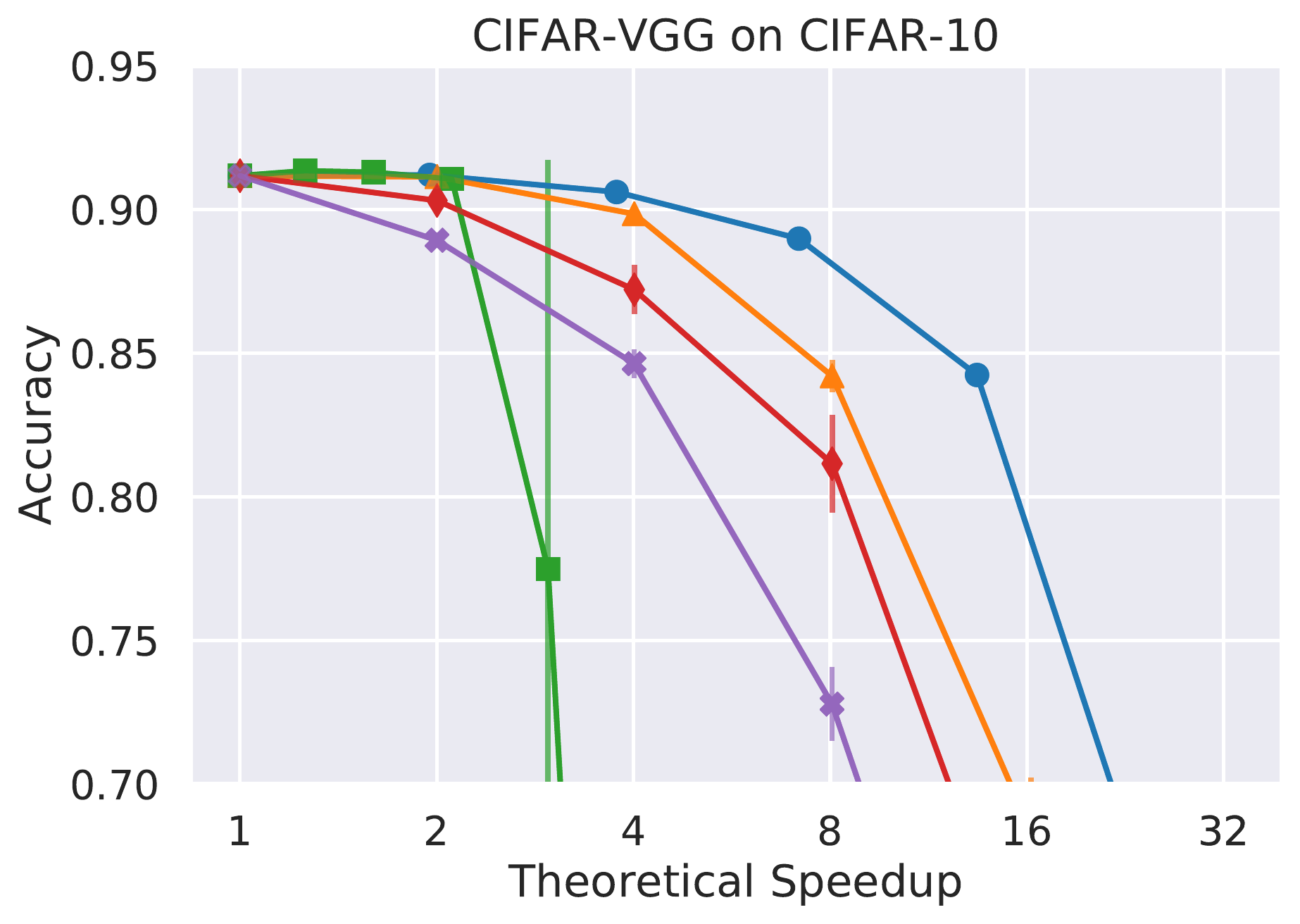}
\caption{Accuracy vs theoretical speedup for CIFAR-VGG on CIFAR-10}
\end{minipage}
\end{figure*}

\begin{figure*}
\begin{minipage}[b]{.45\textwidth}
\centering
\includegraphics[width=\linewidth]{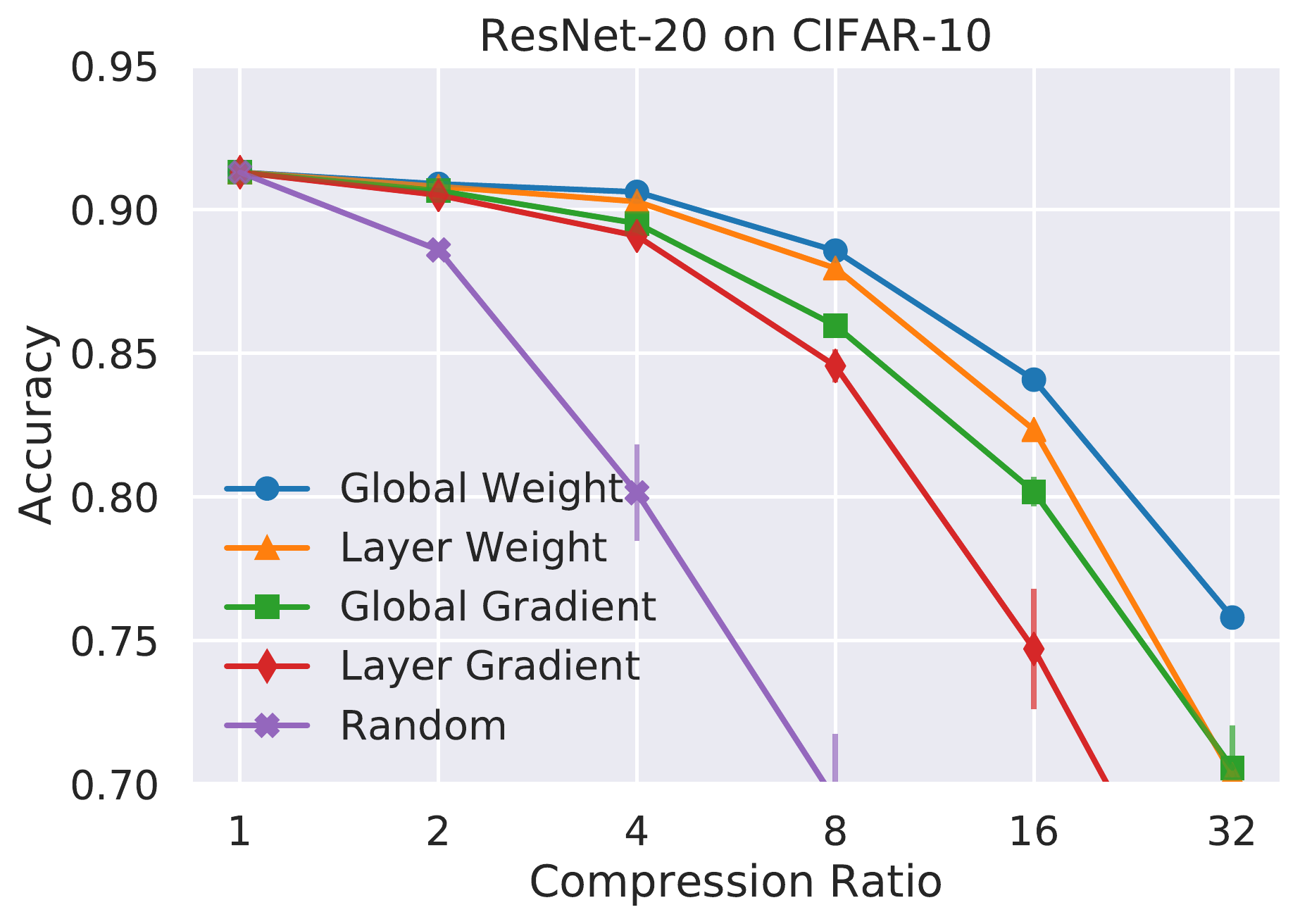}
\caption{Accuracy for several levels of compression for ResNet-20 on CIFAR-10}
\end{minipage}
\hfill
\begin{minipage}[b]{.45\textwidth}
\centering
\includegraphics[width=\linewidth]{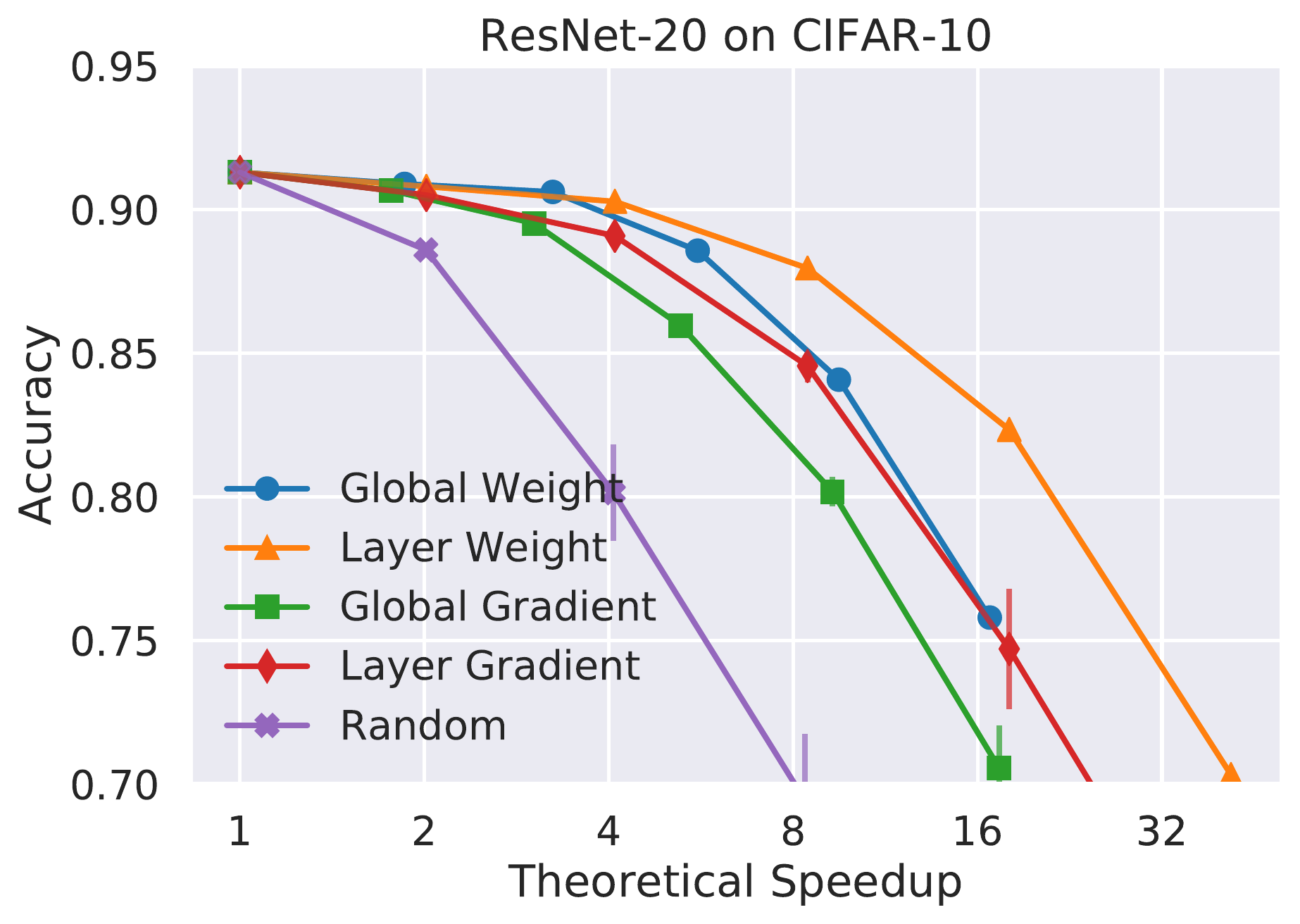}
\caption{Accuracy vs theoretical speedup for ResNet-20 on CIFAR-10}
\end{minipage}
\end{figure*}

\begin{figure*}
\begin{minipage}[b]{.45\textwidth}
\centering
\includegraphics[width=\linewidth]{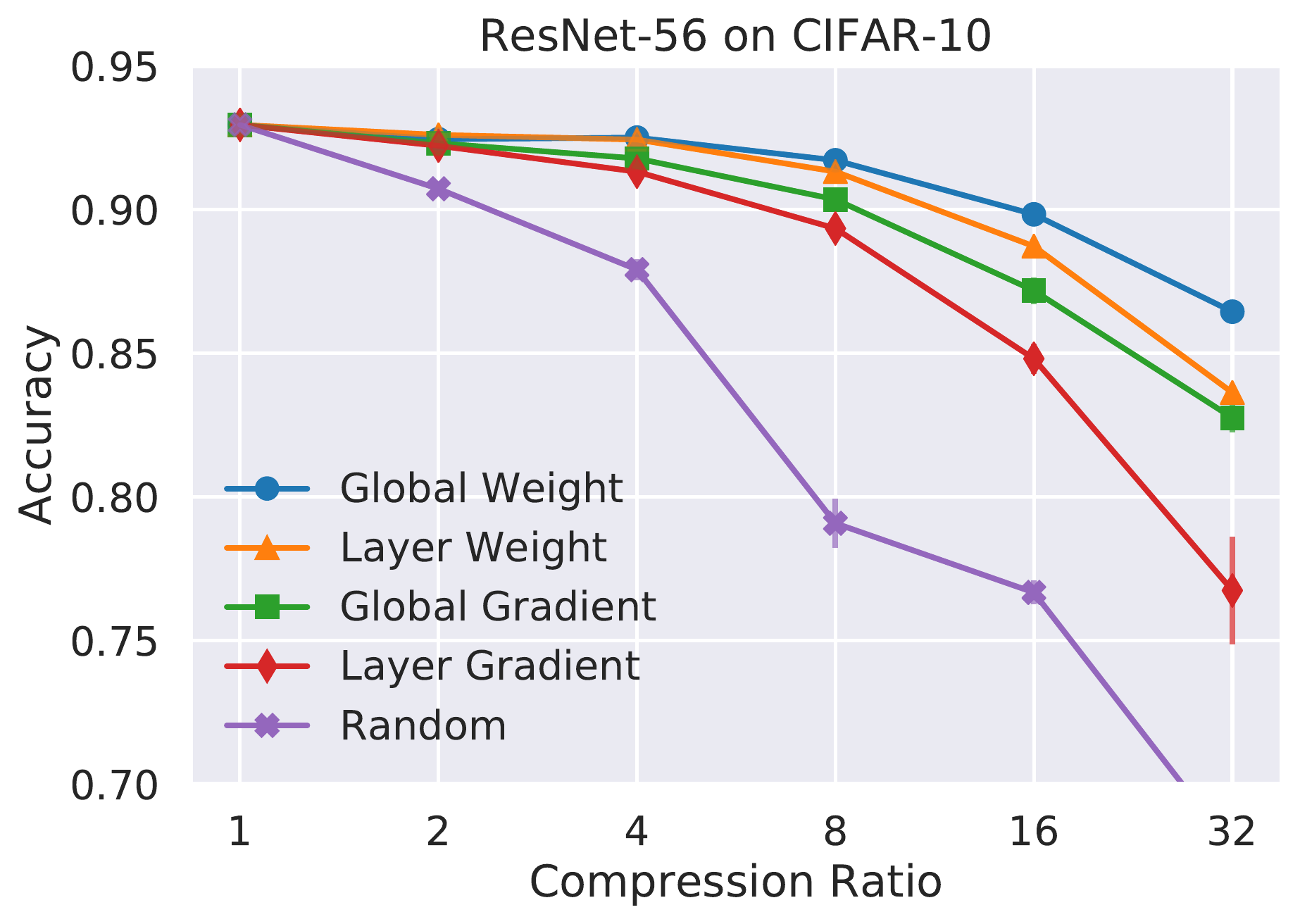}
\captionof{figure}{Accuracy for several levels of compression for ResNet-56 on CIFAR-10}
\end{minipage}
\hfill
\begin{minipage}[b]{.45\textwidth}
\centering
\includegraphics[width=\linewidth]{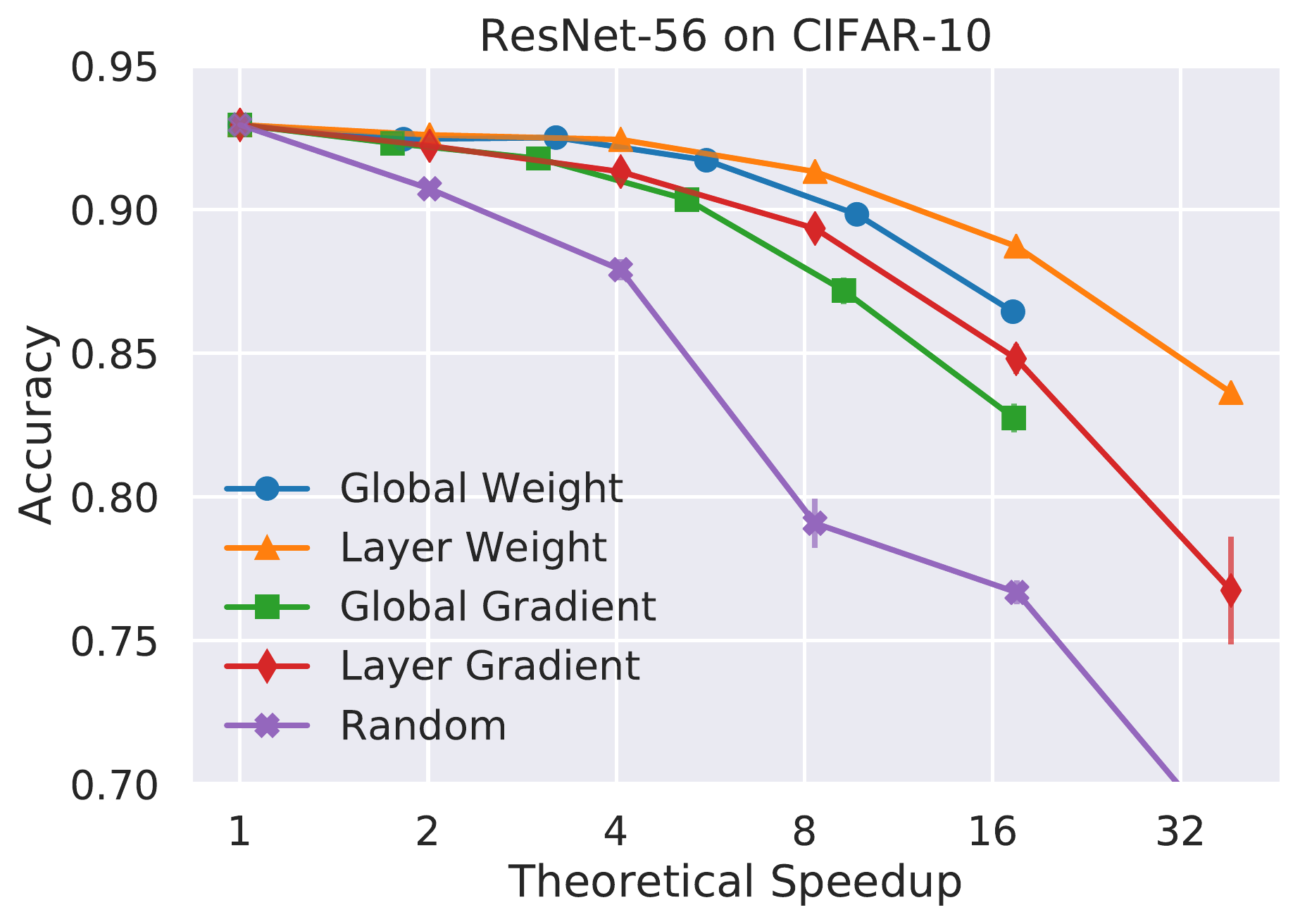}
\caption{Accuracy vs theoretical speedup for ResNet-56 on CIFAR-10}
\end{minipage}
\end{figure*}


\begin{figure*}
\begin{minipage}[b]{.45\textwidth}
\centering
\includegraphics[width=\linewidth]{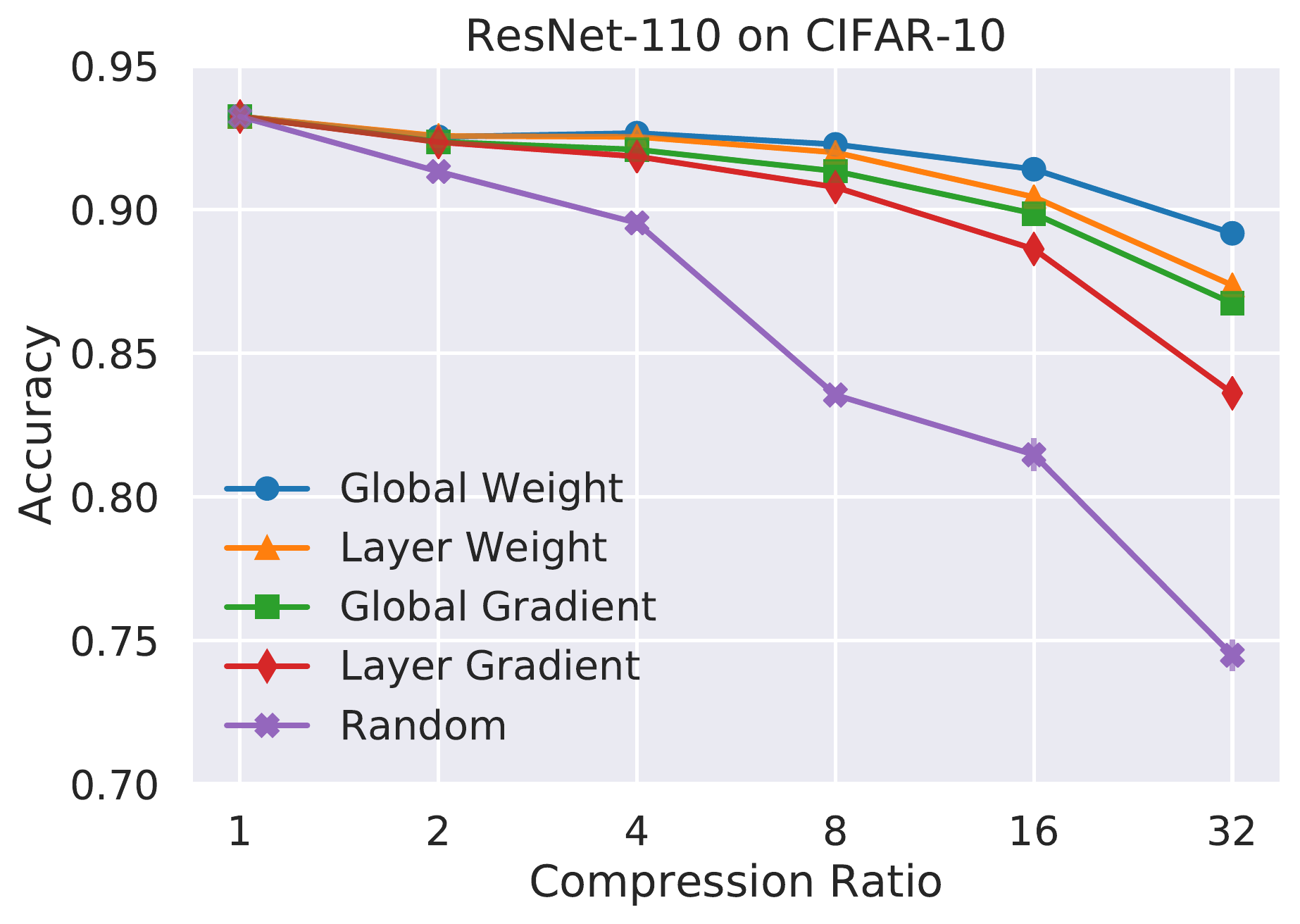}
\caption{Accuracy for several levels of compression for ResNet-110 on CIFAR-10}
\end{minipage}
\hfill
\begin{minipage}[b]{.45\textwidth}
\centering
\includegraphics[width=\linewidth]{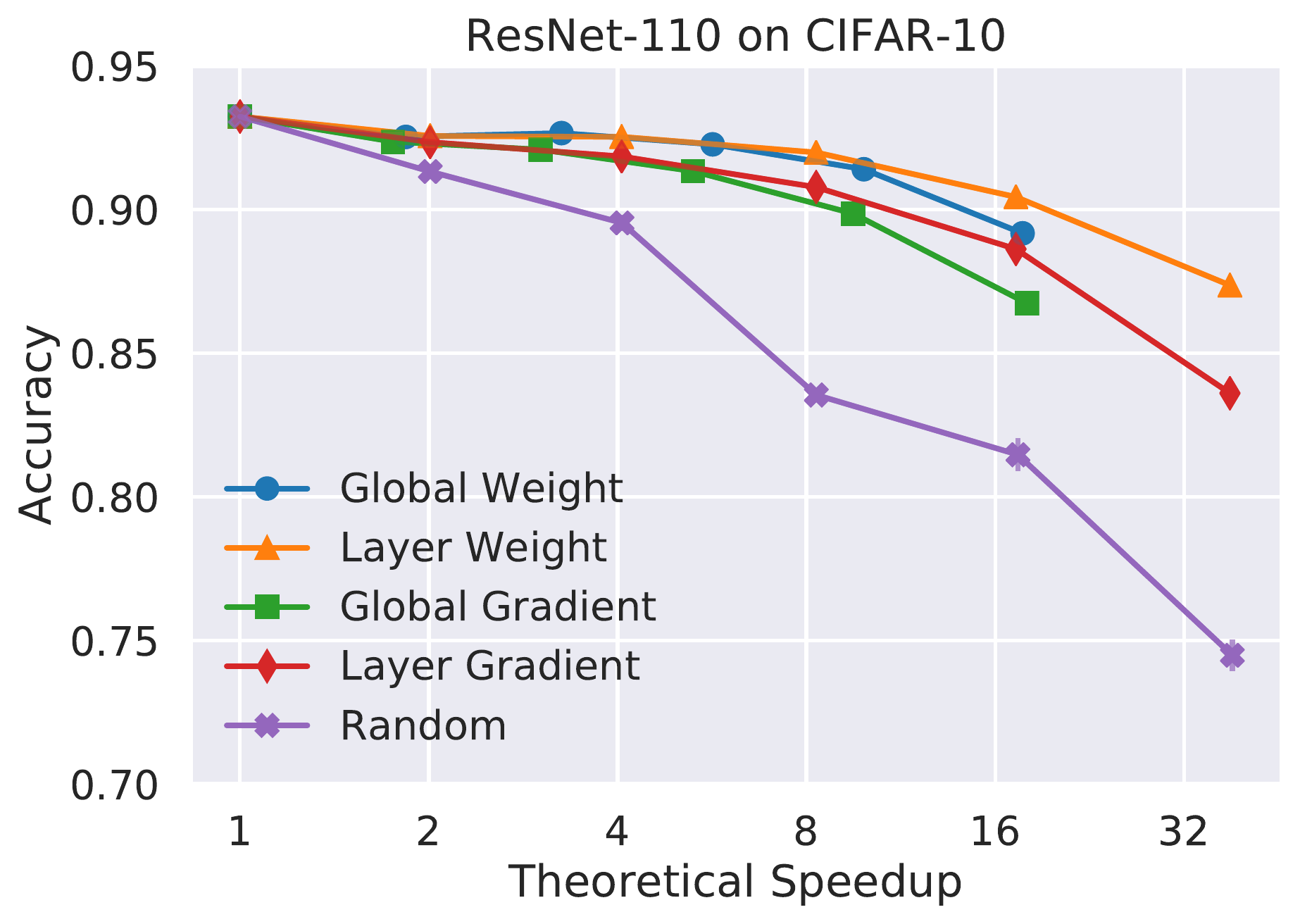}
\caption{Accuracy vs theoretical speedup for ResNet-110 on CIFAR-10}
\end{minipage}
\end{figure*}

\begin{figure*}
\begin{minipage}[b]{.45\textwidth}
\centering
\includegraphics[width=\linewidth]{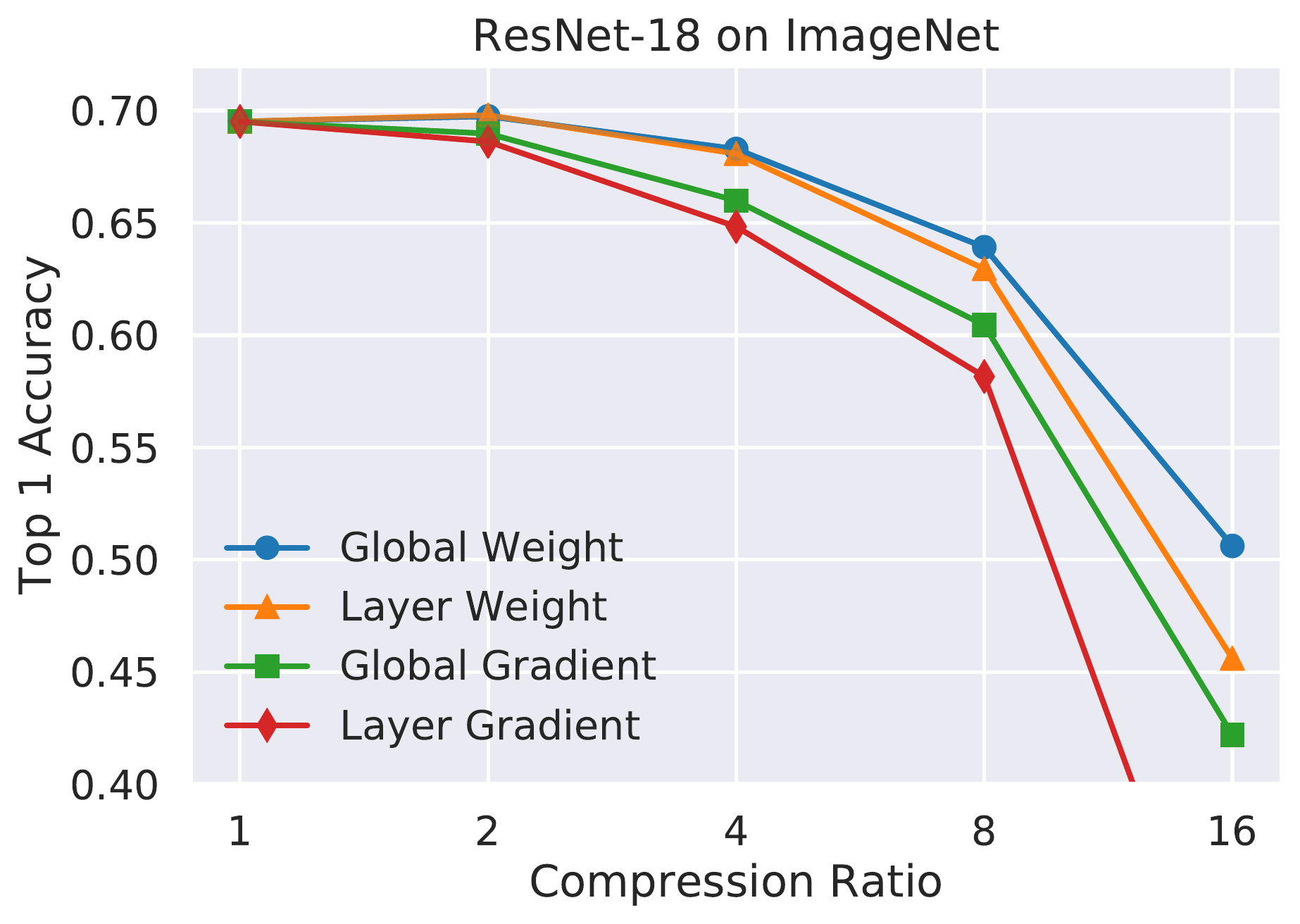}
\caption{Accuracy for several levels of compression for ResNet-18 on ImageNet}
\end{minipage}
\hfill
\begin{minipage}[b]{.45\textwidth}
\centering
\includegraphics[width=\linewidth]{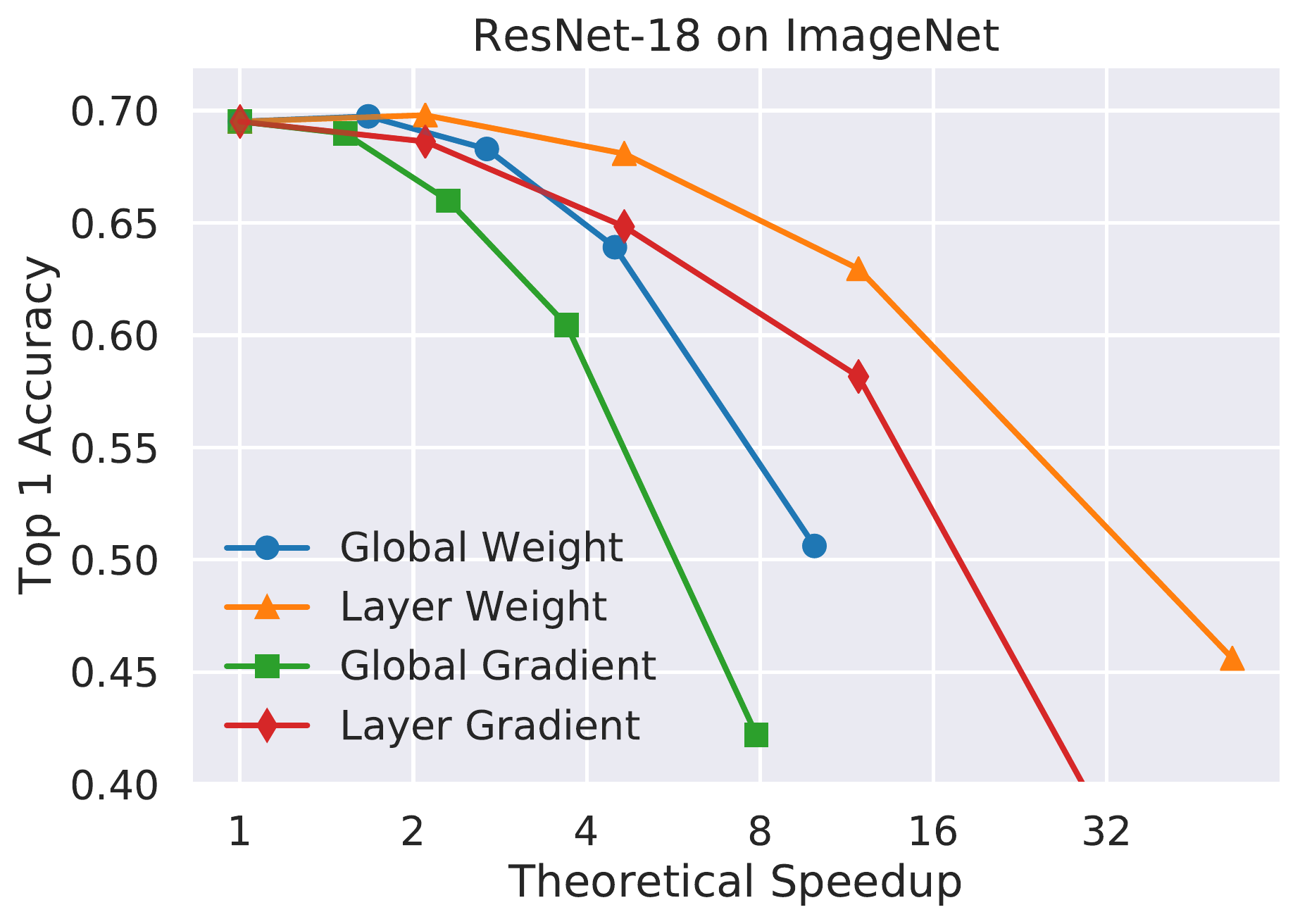}
\caption{Accuracy vs theoretical speedup for ResNet-18 on ImageNet}
\end{minipage}
\end{figure*}